\def\paperID{2396}
\def\confName{CVPR}
\def\confYear{2024}

\def\paperTitle{A General and Efficient Training for Transformer via Token Expansion}

\def\authorBlock{
    Wenxuan Huang$^1$\thanks{Equal contribution.\qquad\qquad\textsuperscript{\Letter}Corresponding author.} \qquad
    Yunhang Shen$^2$\footnotemark[1] \qquad
    Jiao Xie$^3$ \qquad
    Baochang Zhang$^4$ \qquad
    Gaoqi He$^1$ \\
    Ke Li$^2$ \qquad
    Xing Sun$^2$ \qquad
    Shaohui Lin$^{1,5}$\textsuperscript{\Letter} \\
    {\normalsize$^1$School of Computer Science and
Technology, East China Normal University, Shanghai, China} \\
    {\normalsize$^2$Tencent Youtu Lab, China} \qquad
    {\normalsize$^3$Xiamen University, China} \qquad
    {\normalsize$^4$Beihang University, China} \\
    {\normalsize$^5$Key Laboratory of Advanced Theory and Application in Statistics and Data Science - MOE, China} \\
    {\tt\small osilly0616@gmail.com, shenyunhang01@gmail.com, jiaoxie1990@126.com, bczhang@buaa.edu.cn} \\
    {\tt\small gqhe@cs.ecnu.edu.cn, tristanli.sh@gmail.com, winfred.sun@gmail.com, shaohuilin007@gmail.com}
}

\def\OurMethodFullName{Token Expansion\xspace}
\def\OurMethod{ToE\xspace}

\newif\ifreview 
\newif\ifarxiv \newcommand{\arxiv}{\arxivtrue}
\newif\ifcamera 
\newif\ifrebuttal 

\arxiv

\pdfoutput=1
\documentclass[10pt,twocolumn,letterpaper]{article}
\ifreview \usepackage[review]{cvpr} \fi
\ifarxiv \usepackage[pagenumbers]{cvpr} \fi
\ifrebuttal \usepackage[rebuttal]{cvpr} \fi
\ifcamera \usepackage{cvpr} \fi

\usepackage{graphicx}
\usepackage{amsmath}
\usepackage{amssymb}
\usepackage{booktabs}
\usepackage{times}
\usepackage{microtype}
\usepackage{epsfig}
\usepackage[table,xcdraw,dvipsnames]{xcolor}
\usepackage{caption}
\usepackage{float}
\usepackage{placeins}
\usepackage{color, colortbl}
\usepackage{stfloats}
\usepackage{enumitem}
\usepackage{tabularx}
\usepackage{xstring}
\usepackage{multirow}
\usepackage{xspace}
\usepackage{url}
\usepackage{subcaption}
\usepackage{xcolor}
\usepackage[hang,flushmargin]{footmisc}

\usepackage[ruled,linesnumbered]{algorithm2e}

\usepackage{threeparttable}

\usepackage{lipsum}

\usepackage{marvosym}

\ifcamera \usepackage[accsupp]{axessibility} \fi

\newcommand{\nbf}[1]{{\noindent \textbf{#1.}}}

\ifarxiv  \fi

\newcommand{\R}[1]{{%
            \textbf{%
                \ifstrequal{#1}{1}{\textcolor{red}{R#1}}{%
                    \ifstrequal{#1}{2}{\textcolor{blue}{R#1}}{%
                        \ifstrequal{#1}{3}{\textcolor{magenta}{R#1}}{%
                            \ifstrequal{#1}{4}{\textcolor{teal}{R#1}}{%
                                \textcolor{cyan}{R#1}%
                            }}}}%
            }%
        }}

\usepackage{xr-hyper}

\makeatletter
\newcommand*{\addFileDependency}[1]{
  \typeout{(#1)}
  \@addtofilelist{#1}
  \IfFileExists{#1}{}{\typeout{No file #1.}}
}

\makeatother

\definecolor{cvprblue}{rgb}{0.21,0.49,0.74}
\usepackage[pagebackref,breaklinks,colorlinks,citecolor=cvprblue]{hyperref}
\usepackage[capitalize]{cleveref}
\crefname{section}{Sec.}{Secs.}
\crefname{table}{Table}{Tables}
\crefname{figure}{Fig.}{Figs.}

\usepackage[normalem]{ulem}

\begin{document}
\title{\paperTitle}
\author{\authorBlock}
\maketitle

\begin{abstract}
The remarkable performance of Vision Transformers~(ViTs) typically requires an extremely large training cost. Existing methods have attempted to accelerate the training of ViTs, yet typically disregard method universality with accuracy dropping.
Meanwhile, they break the training consistency of the original transformers, including the consistency of hyper-parameters, architecture, and strategy, which prevents them from being widely applied to different Transformer networks. In this paper, we propose a novel token growth scheme Token Expansion (termed ToE) to achieve consistent training acceleration for ViTs. We introduce an ``initialization-expansion-merging'' pipeline to maintain the integrity of the intermediate feature distribution of original transformers, preventing the loss of crucial learnable information in the training process. ToE can not only be seamlessly integrated into the training and fine-tuning process of transformers (\eg, DeiT and LV-ViT), but also effective for efficient training frameworks (\eg, EfficientTrain), without twisting the original training hyper-parameters, architecture, and introducing additional training strategies. Extensive experiments demonstrate that ToE achieves about $1.3\times$ faster for the training of ViTs in a lossless manner, or even with performance gains over the full-token training baselines.
Code is available at \url{https://github.com/Osilly/TokenExpansion}.
\end{abstract}

\section{Introduction}
\label{sec:intro}

Transformers have achieved excellent performance in the tasks of natural language processing~(NLP)~\cite{vaswani2017attention, kenton2019bert, brown2020language} and computer vision~\cite{touvron2021training, jiang2021all, carion2020end, xie2021segformer}.
Despite their great success, modern Transformer models typically require extremely large parameters and computation consumption due to the quadratic computational complexity in the self-attention module.
For example, ViT-H/14~\cite{dosovitskiy2020image} requires $\sim$1,000B FLOPs, which is $250\times$ larger than ResNet-50~\cite{he2016deep}.
The entire training process needs a significant amount of computing resources to reach model convergence, resulting in a substantial computation overhead.
To reduce the computational cost of large models, there has been growing research attention on accelerating Transformers for either training or inference.

\begin{table}[t]
\caption{Training results for DeiT~\cite{touvron2021training} on ImageNet-1K.
DeiT does \textit{not} use the EMA strategy by default. a/b in the column of Top-1 Acc. means without/with EMA strategy using the official GitHub repo.
The training time is averagely measured on one/four NVIDIA RTX A6000 GPUs $3$ times with a batch size of $1,024$ for DeiT-Tiny/Base, respectively.}
\centering
\resizebox{0.99\linewidth}{!}{
    \begin{tabular}{c|c|ccc|cccc}
        \toprule
        \multirow{2}{*}{Model}     & \multirow{2}{*}{Method}                                                   & \multicolumn{3}{c|}{Training consistency}  & \multirow{2}{*}{Top-1 Acc. (\%)}  & \multicolumn{1}{c}{Training time}                                                                              \\
        \cmidrule{3-5}
                                   &                                                                           & Hyper                            & Arch                     & Strategy                         &                                           & (GPU hours)           \\
        \midrule
        \multirow{5}{*}{Tiny} & Baseline~\cite{touvron2021training}                                       & $-$                               & $-$                               & $-$                               & 72.2                                      &  54.6h                           \\
                                   &S$^2$ViTE (600 epoch)~\cite{chen2021chasing}            & $\times$                          & $\times$                          & $\checkmark$                      & 70.1 (-2.1)                               &  $-$                           \\
                                   & ToMe$_{r_{8}\rightarrow}^{\mathrm{DeiT}}$~\cite{bolya2022token} & $\checkmark$                      & $\times$                          & $\checkmark$                      & 71.7 (-0.5)                               &  53.3h                           \\
                                   & NetworkExpansion$_{6\rightarrow 12}$~\cite{ding2023network}     & $\checkmark$                      & $\checkmark$                      & $\times$                          & 70.3 (-1.9) / 70.1 (-2.1)                 &  43.2h                           \\
                                   & \cellcolor[gray]{0.8}ToE$_{r_1=0.5}$ (Ours)                               & \cellcolor[gray]{0.8}$\checkmark$ & \cellcolor[gray]{0.8}$\checkmark$ & \cellcolor[gray]{0.8}$\checkmark$ & \cellcolor[gray]{0.8}\textbf{72.6 (+0.4)} & \cellcolor[gray]{0.8}44.2h       \\
        \midrule
        \multirow{4}{*}{Base} & Baseline~\cite{touvron2021training}                                       & $-$                               & $-$                               & $-$                               & 81.8                                      & 292.8h                      \\
                                   & StackBERT~\cite{gong2019efficient}                              & $\checkmark$                      & $\checkmark$                      & $\times$                          & 80.8 (-1.0)                               & 231.6h                      \\
                                   & NetworkExpansion$_{6\rightarrow 12}$~\cite{ding2023network}     & $\checkmark$                      & $\checkmark$                      & $\times$                          & 81.0 (-0.8) / 81.5 (-0.3)                 & 226.8h                      \\
                                   & \cellcolor[gray]{0.8}ToE$_{r_1=0.5}$ (Ours)                               & \cellcolor[gray]{0.8}$\checkmark$ & \cellcolor[gray]{0.8}$\checkmark$ & \cellcolor[gray]{0.8}$\checkmark$ & \cellcolor[gray]{0.8}\textbf{81.6 (-0.2)} & \cellcolor[gray]{0.8}231.2h \\
        \bottomrule
    \end{tabular}

}
\label{tab:consistency}
\end{table}

Existing Transformer pruning methods~\cite{yang2023global,yu2022width,lagunas2021block,xia2022structured,rao2021dynamicvit,meng2022adavit,fayyaz2022adaptive,kong2022spvit,yin2022vit} aim to reduce the inference complexity.
Among them, structure pruning~\cite{yang2023global, yu2022width, lagunas2021block, xia2022structured} and token pruning~\cite{rao2021dynamicvit, meng2022adavit, fayyaz2022adaptive, kong2022spvit, yin2022vit} focus on reducing the neurons or tokens of Transformers to accelerate the inference.
However, these pruning methods require additional training computational cost %
in each forward-backward iteration to determine which neurons or tokens are important enough to be retained, or the fine-tuning for pruned models.
Recently, Transformer quantization~\cite{xu2023q,li2022q,he2023bivit,le2023binaryvit} accelerates the inference via low-bit computation, but they also cannot reduce the training computation cost.
Thus, it is challenging for them to effectively accelerate the training of Transformers in practical scenarios, \eg, cloud service.

To reduce the training computation overhead, recent works~\cite{gong2019efficient,  chen2022bert2bert, ding2023network, yuan2020growing, wen2020autogrow} have proposed structure growth methods.
They update a smaller number of model parameters during the early stages of training and gradually increase the number of parameters involved in the updating process as training progresses.
However, the existing methods fail to achieve general transformer training acceleration without accuracy dropping (shown in Tab.~\ref{tab:consistency}), and they break the \textit{training consistency} of the original transformers from three perspectives:
(1) Hyper-parameter consistency. 
Existing methods (\eg, SViTE~\cite{chen2021chasing}) delicately tune training hyper-parameters (\emph{e.g.}, learning rate and epoch number) of the original models, which are sensitive to individual ViTs~\cite{touvron2021training} and require additional trial-and-error costs for different networks.
(2) Architecture consistency. 
Existing methods~\cite{chen2021chasing, bolya2022token} alter the final model architectures, which may deviate from the user's requirements and potentially necessitates additional hardware/software support to implement real training speedup.
For example, ToMe~\cite{bolya2022token} progressively merges similar tokens layer-by-layer to reduce the number of tokens in ViTs during training, which replaces the attention operators with the weighted average attention modules, generating a different model architecture that deviates from the original Transformer. Moreover, it cannot significantly accelerate the practical training due to the unfriendly computation.
(3) Strategy consistency.
Existing methods~\cite{gong2019efficient, chen2022bert2bert, ding2023network} may suffer from performance deterioration across different Transformers by adding additional training strategies, such as EMA and reset optimizer states.
It means the effectiveness of these strategies is for specific models, which limits the method's universality whether employing them for training.
In Tab.~\ref{tab:consistency}, the extra EMA strategy in ~\cite{ding2023network} plays different roles to the performance across different models, \emph{i.e.}, the effectiveness for DeiT-base but not for DeiT-tiny. 
Thus, this begs our rethinkings: \emph{How to implement real and friendly training speedup for Transformers while keeping the training consistency and high accuracy?}

\begin{figure*}[t]
    \centering
    \includegraphics[scale = 0.31]{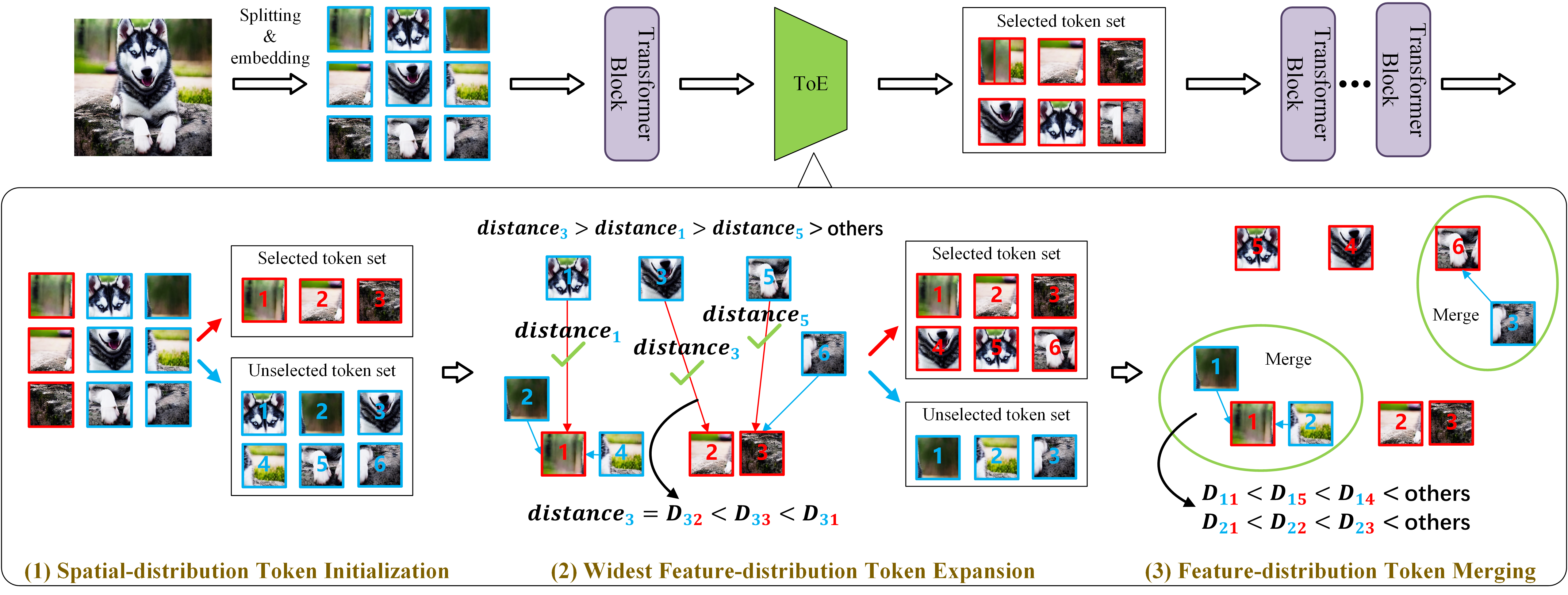}
    \caption{
    The ``initialization-expansion-merging'' pipeline of proposed ToE.
    We take the $1$st training stage ($\delta=1$), the kept rate $r_1=2r_0=\frac{2}{3}$, the repetition step $k=1$ as example.
    ToE is only added after the first Transformer block to guide the token selection and usage.
    During training, steps (1), (2), and (3) are performed for each iteration with the reduction of token numbers.
    First, seed tokens are selected for token initialization through step (1).
    Then, the number of tokens is expanded via step (2) for token expansion.
    Finally, we merge the unselected token set (blue boxes) into the selected one (red boxes) with the close feature distributions in step (3) for token merging.
    During testing, ToE can be safely removed to generate the same Transformer architecture as the original full-token Transformer.
    }
    \label{fig:pipeline}
\end{figure*}

To answer the above question, we propose a novel token growth scheme, \textit{Token Expansion} (termed \textbf{\OurMethod}) to achieve general training acceleration for ViTs, while adhering to the training consistency of original models.
Specifically, we present an ``initialization-expansion-merging'' pipeline (in Fig.~\ref{fig:pipeline}) to maintain the integrity of the intermediate feature distribution of original transformers, preventing the loss of crucial learnable information during the accelerated training process.
Similar to structure growth methods, we initially involve a limited number of tokens to participate in training and gradually grow the token number during training progress, eventually reaching the utilization of the entire token set.
Then, a \textit{widest feature-distribution token expansion} is introduced to make the feature distributions of the selected token set as wide as possible.
Additionally, a \textit{feature-distribution token merging} combines the tokens with close feature distributions to further avoid information loss.
ToE not only accelerates the training and fine-tuning process of popular Transformers in a lossless manner or even with performance improvement, but also can be integrated into the existing efficient training frameworks (\eg, EfficientTrain~\cite{wang2023efficienttrain}) for further performance improvement, without twisting the original training hyper-parameters, architecture, and introducing additional training strategies.

Our main contributions can be summarized as follows:
\begin{itemize}
      \item
            We propose \OurMethod, a novel token growth scheme to accelerate ViTs from the perspective of tokens.
            \OurMethod is a consistent training acceleration method and can be seamlessly integrated into the training and fine-tuning process of transformers without any modifications to the original training hyper-parameters, architecture, and strategies.
      \item
            We %
            propose an effective ``initialization-expansion-merging'' framework to avoid the token information loss by maintaining the integrity of the intermediate feature distribution. 
      \item
            Extensive experiments demonstrate that \OurMethod accelerates the training and fine-tuning process of ViTs with a negligible accuracy drop or even surpassing the original full-token counterparts, which outperforms previous SOTA methods.
\end{itemize}

\section{Related Work}
\label{sec:related}

\subsection{Training Acceleration for Transformers}
As mentioned above, many existing works focus on accelerating the training of transformers from the perspective of structural parameters.
These structure methods~\cite{gong2019efficient, chen2022bert2bert, ding2023network, chen2021chasing, li2022automated, pan2022budgeted} reduce the number of updated parameters in the training process to save the computational cost.
In contrast, the proposed \OurMethod accelerates training from the perspective of reducing token redundancy.
In other words, \OurMethod computes a smaller number of tokens but still optimizes all parameters.
It avoids potential performance drops in many structure growth methods due to the inconsistent structures of before-and-after models during structure growth and resetting of optimizer state when updating new structural parameters. %

ToMe~\cite{bolya2022token} uses a limited number of tokens to participate in training and progressively merges similar tokens layer-by-layer, which changes the attention operator in inference.
\OurMethod also involves merging tokens with close feature distributions by \textit{feature-distribution token merging}.
However, our merging strategy is performed only once at the end of the ``initialization-expansion-merging'' pipeline during training, which prevents the information loss of tokens.
This ensures that \OurMethod avoids the mismatch between practical and theoretical acceleration caused by excessive merging operations and operator modifications. %

Additionally, several works~\cite{lee2022deduplicating, wang2023efficienttrain, tan2021efficientnetv2, mcdanel2022accelerating} also consider to reduce the data for training. %
The work in~\cite{lee2022deduplicating} deduplicates training datasets to save computational resources.
Unfortunately, it usually introduces additional computational costs and sometimes becomes a bottleneck by using additional time to process datasets during training~\cite{shen2023efficient}.
PSS~\cite{mcdanel2022accelerating} uses fewer patches obtained by splitting images during training.
EfficientTrain~\cite{wang2023efficienttrain} and PL~\cite{tan2021efficientnetv2} use images of different sizes and additional data augmentation.
However, EfficientTrain and PL change the training pipelines that differ from the training of the original model, \eg, hyper-parameters.
Moreover, the above methods consider the properties of training data.
In contrast, \OurMethod focuses on the crucial learnable information in the intermediate feature space of transformers.
Thus, \OurMethod can be integrated into the above methods in a plug-and-play manner to further enhance training efficiency. %

\subsection{Training Acceleration for CNNs}
Prior efficient training acceleration methods have explored ways to speed up the training of CNN models~\cite{yang2023efficient, ye2020accelerating, fu2020fractrain, wang2019e2, li2019budgeted, zhang2019autoassist}. For example,
works in~\cite{yang2023efficient, ye2020accelerating} consider pruning gradients to reduce training computation costs.
Works in~\cite{fu2020fractrain, wang2019e2} attempt to use quantization technical to achieve training acceleration.
Others try to reduce training time either by reducing the number of optimization iterations with a linear decay for the learning rate~\cite{li2019budgeted} or skipping easy samples that contribute little to loss reduction~\cite{zhang2019autoassist}.
However, these methods may not be directly applied to Transformers for training acceleration due to the specific architectural differences between transformers and CNNs.
Differently, \OurMethod focuses on the training acceleration for Transformers on the token dimension.

\subsection{Transformer pruning}
Transformer pruning methods typically reduce parameters or tokens to generate sparse Transformers for fast inference.
Structure pruning methods~\cite{yang2023global,yu2022width,lagunas2021block,xia2022structured} attempted to prune the structures of transformers. Token pruning methods~\cite{rao2021dynamicvit, meng2022adavit, fayyaz2022adaptive, kong2022spvit, yin2022vit} focused on dynamically determining the importance of input tokens and pruning them during inference.

The key differences between our method and transformer pruning methods are two-fold.
(1) Transformer pruning methods primarily aim to accelerate transformer inference, while our target is for training acceleration.
(2) We obtain a dense model after training by token growth, which is entirely consistent with the original model for inference. In contrast, pruning methods generate sparse models after training.

\section{Method}
\label{sec:method}

\subsection{Preliminaries and Notations}
\label{subsec:preliminaries}

Given a Transformer with $L$ blocks, we denote the sets of input and output tokens for the $l$-th block as $\mathcal{S}_{l-1}$ and $\mathcal{S}_{l}$ with $l\in \{1, 2, \cdots, L\}$, respectively.
The index set of output tokens for the $l$-th block is defined as $\mathcal{I} = \{1, 2, \cdots, N_l\}$, where $N_l$ is the number of output tokens for the $l$-th block.
We further denote the $i$-th token of the output tokens for the $l$-th block as $\mit{t}_{l,i}\in \mathbb{R}^{d}$, thus $\mathcal{S}_{l} = \{\mit{t}_{l,i}|\forall i\in\mathcal{I}\}$.

For the $l$-th Transformer block, we consider to reduce the output tokens to a specified size $N_l^{'} = \lfloor r N_l\rfloor$, where $r\in \left( 0,1 \right]$ is the kept rate of tokens, and $\lfloor\cdot\rfloor$ is a floor function.
Further, we define the index set of kept tokens as $\mathcal{I}^{'} = \{1, 2, \cdots, N_l^{'}\}$ and we obtain a subset $\mathcal{S}_{l}^{'} = \{\mit{t}_{l,i}^{'} | \forall i \in \mathcal{I}^{'}\}$ of output tokens.
When the output tokens of the $l$-th block are reduced, this results in a corresponding reduction in the quantity of input tokens for blocks beyond the $l$-th block.
Furthermore, the computational complexity of self-attention blocks and MLP layers in Transformers is directly proportional to the number of input tokens. 
According to the work~\cite{epoch2021backwardforwardFLOPratio}, the computation in the forward and backward propagation of modern neural networks roughly conforms to 1:2.
Therefore,  the reduction of tokens significantly accelerates the computation in both the forward and backward propagations during training if $r < 1$. Note that, to reduce the complex search computation for the kept rate of tokens $r$ across all Transformer blocks, we simply and effectively set $r$ to be the same in all blocks that benefit from acceleration.

\subsection{Overview of ToE}
\label{subsec:token growth strategy}

As shown in Fig.~\ref{fig:pipeline}, \OurMethod initially selects a significantly small number of tokens, then progressively grows to the final full-token same as the original Transformer, thereby achieving training acceleration.
We divide the origin training process into $N_g$ stages on average.
We use a limited number of tokens to participate in each training stage and gradually grow the token number along with the training stages.
The token growth strategy consists of three steps:

\textbf{(1) Initial token selection as the seed tokens}.
we initially select $\lfloor r_0 N_l\rfloor$ output tokens from the origin token set $\mathcal{S}_{l}$ as the seed token set by using Uniform sampling on the index set $\mathcal{I}$, where $r_0$ represents the pre-defined initial kept rate, which is default set to less than 0.3 in our experiments unless otherwise specified.

\textbf{(2) Token expansion}.
In the $\delta$-th ($\delta\in \{1, 2, \cdots, N_g\}$) training stage, we perform $\delta$ times token expansion to preserve the integrity of the original intermediate feature space.
Furthermore, we pre-define the keep rate of the first stage to be $r_1$.
The kept rate of $\delta$-th stage $r_\delta$ is computed as:
\begin{equation}
\small
    \begin{split}
        & \mu_{\delta} =
        \begin{cases}
            r_1 - r_0,       & \text{if }\delta=1, \\
            \frac{ 1 - r_1 }{ N_g - 1 }, & \text{otherwise},
        \end{cases} \\
        & r_\delta = r_{\delta-1} + \mu_{\delta}, \\
    \end{split}
    \label{eq1}
\end{equation}
where $\mu_{\delta}$ is the token expansion rate in the $\delta$-th training stage and $r_1 = 2 \cdot r_0\in \left( 0,1 \right]$.
After the $\delta$ times token expansion, we select $\lfloor r_\delta N_l\rfloor$ tokens from the full-token set $\mathcal{S}_{l}$. In Sec.~\ref{sssec:wfdte}, we will introduce the widest feature-distribution token expansion method to select $\lfloor r_\delta N_l\rfloor$ tokens, which aims to expand the token distribution space to effectively present full-token feature distribution.

\textbf{(3) Token merging}.
To further avoid information loss during the training process, we consider merging the unselected tokens into the selected ones in the token expansion process, which retains effective information of the unselected tokens in the merged token set $\mathcal{S}_{l}^{'}$.
Inspired by ToMe~\cite{bolya2022token}, we merge \textit{averagely} the tokens that the feature distributions are close as one new token, which is further introduced in Sec.~\ref{sssec:fdtm}.

During training, \OurMethod performs steps (1), (2), and (3) on the original full-token set for each training iteration, which reduces the number of tokens involved in training while retaining the effective information from the full-token set.

\subsection{Token Expansion}
\label{subsec:token expansion}
In this Section, we introduce the proposed \OurMethod method, including spatial-distribution token initialization, widest feature-distribution token expansion, feature-distribution token merging, and its optimization.

\subsubsection{Spatial-distribution Token Initialization}
\label{sssec:sdti}
For the initialization, we apply a simple strategy to select the initial token set from $\mathcal{S}_l$.
We define the index of the initial token set as:
\begin{equation}
\small
    \mathcal{I}^{(I)} = \{i|\forall i\bmod \lfloor\frac{1}{r_0}\rfloor=1 \land \forall i\in\mathcal{I}\}.
    \label{eq2}
\end{equation}
The selected token set and the unselected tokens set can be expressed as $\mathbb{A}=\{\mit{t}_{l,i}|\forall i\in\mathcal{I}^{(I)}\}$ and $\mathbb{B}=\mathcal{S}_l-\mathbb{A}$, respectively.
This initialization selection strategy is based on spatial distribution.
It indicates that we choose one token out of every $\lfloor\frac{1}{r_0}\rfloor$ tokens from the original token set and add it to the initial token set.
Our strategy is simple, yet effective, to ensure that the initially selected tokens provide broad spatial coverage across the image patches.

\subsubsection{Widest Feature-distribution Token Expansion}
\label{sssec:wfdte}
Previous works~\cite{bolya2022token, rao2021dynamicvit} show that the intermediate feature space in modern Transformers is \textit{overparameterized}, such that they prune the full-token Transformers to be sparse ones. Actually, through the above token initialization, we obtain the sparse Transformers. However, the performance drops significantly if we only train on these selected tokens. Thus, we consider to grow the number of tokens, which is expected to preserve the integrity of the original intermediate feature space and avoid the loss of tokens containing valuable information.
Inspired by this, we seek to maintain the integrity of the intermediate feature distribution.
Intuitively, when the feature distributions of two token sets are sufficiently close, they have similar information that can be used to effectively represent each other.
In contrast, given one token whose feature distribution deviates significantly from all other tokens in the token set, it will be difficult to be adequately represented by other tokens, such that we expect to select this token to underscore its importance in the token expansion.

To this end, we propose the widest feature-distribution token expansion strategy.
Specifically, we perform the expanding operation on the selected tokens from the initialized set.
For the $\delta$-th stage of token expansion, we consider the selected token set $\mathbb{A}\in\mathbb{R}^{|\mathbb{A}|\times d}$ and the unselected token set $\mathbb{B}\in\mathbb{R}^{|\mathbb{B}|\times d}$ as the 2D matrices, where $|\cdot|$ and $d$ respectively denote the number of tokens and feature dimension, and $|\mathbb{A}|+|\mathbb{B}|=N_l$. 
We utilize \textit{Cosine Distance} as the metric to measure the distance between feature distribution of tokens in these two sets (other metrics see Tab.~\ref{tab:distance}):
\begin{equation}
\small
        \mathcal{D}(\mathbb{B}, \mathbb{A}) = \mathbf{1} - \operatorfont{cos}\left<\mathbb{B}, \mathbb{A}\right> 
        = \mathbf{1} - \frac{\mathbb{B}\mathbb{A}^{\mathrm{T}}}{\left\vert\kern-0.25ex\left\vert\mathbb{B}\right\vert\kern-0.25ex\right\vert \cdot \left\vert\kern-0.25ex\left\vert\mathbb{A}\right\vert\kern-0.25ex\right\vert}, 
    \label{eq3}
\end{equation}
where $\mathbf{1}$ is an all-one matrix.
$\mathcal{D}(\mathbb{B}, \mathbb{A}) \in\mathbb{R}^{|\mathbb{B}|\times |\mathbb{A}|}$ represents the pairwise distances between tokens in $\mathbb{B}$ and $\mathbb{A}$.

We further define the distance between the feature distribution of tokens in $\mathbb{B}$ and its closest token in $\mathbb{A}$ as $distance(\mathbb{B}\rightarrow\mathbb{A}) \in\mathbb{R}^{|\mathbb{B}|}$:
\begin{equation}
\small
    \mathit{distance}(\mathbb{B}\rightarrow\mathbb{A})_i = \text{min}_j(\mathcal{D}(\mathbb{B}, \mathbb{A})_{i,j}),
    \label{eq4}
\end{equation}
where $i\in\{1,\cdots,|B|\}$ and $j\in\{1,\cdots,|A|\}$.
Eq.~\ref{eq4} indicates that we sample the minimal values of the feature-distribution distance matrix $\mathcal{D}(\mathbb{B}, \mathbb{A})$ along the second dimension.
Thus, $\mathit{distance}(\mathbb{B}\rightarrow\mathbb{A})_i$ measures importance of $i$-th token in $\mathbb{B}$.
At this point, we progressively add the most important token to $\mathbb{A}$, which is formulated as:
\begin{equation}
\small
    \begin{array}{c}
        \mathbb{A}=\mathbb{A}+\mit{t}^{*}, \quad
        \mathbb{B}=\mathbb{B}-\mit{t}^{*}, \\
        \mit{t}^{*}=\{\mathbb{B}_{i}|i=\operatorfont{argmax}(\mathit{distance}(\mathbb{B}\rightarrow\mathbb{A}))\},
    \end{array}
    \label{eq5}
\end{equation}
where $\mit{t}^{*}$ is the most important token in $\mathbb{B}$.
When the feature distribution of one token is far from its closest token, it can be said that the feature distribution of this token deviates significantly from that of all other tokens in the token set.
The operation described in Eq.~\ref{eq5} is performed for $\lfloor\mu_{\delta} N_l\rfloor$ times to select $\lfloor\mu_{\delta} N_l\rfloor$ tokens from $\mathbb{B}$ into $\mathbb{A}$.
The widest feature-distribution token expansion strategy ensures that the feature distributions of the selected token set become as wide as possible, preventing the loss of important tokens.
However, as we need to iterate $\lfloor\mu_{\delta} N_l\rfloor$ times expansion, it results in a considerable consumption of computational resources.
Considering the computation parallelization, we modify the expanding operation in Eq.~\ref{eq5} parallelly:
\begin{equation}
\small
    \begin{array}{c}
        \mathbb{A}=\mathbb{A}+\mathcal{S}^{*}, \quad
        \mathbb{B}=\mathbb{B}-\mathcal{S}^{*}, \\
        \mathcal{S}^{*}=\{\mathbb{B}_{i}|i \in \operatorfont{topk}_{\lfloor\mu_{\delta} N_l/{k}\rfloor}(\mathit{distance}(\mathbb{B}\rightarrow\mathbb{A}))\},
    \end{array}
    \label{eq6}
\end{equation}
where $k$ is the pre-defined repetition step of parallel expanding operation, $\mathcal{S}^{*}$ is a token set consisting of the important tokens in $\mathbb{B}$, $\text{topk}_n$ denotes the top argmax with the number of $n$ tokens.
By this way, we only perform $k$ times parallel expanding operation to expand $\lfloor\mu_{\delta} N_l\rfloor$ tokens, and its computational consumption is negligible with small $k$.

\begin{figure*}[t]
    \centering
    \includegraphics[scale = 0.25]{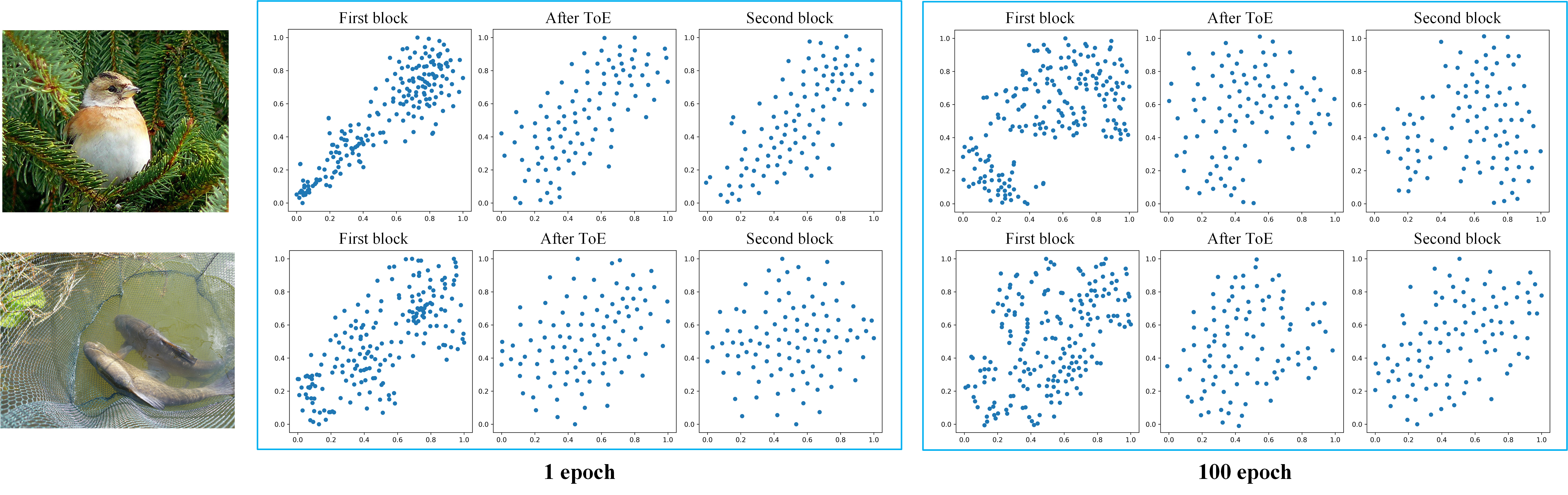}
    \caption{Visualization for the feature distribution of token set.
    We use T-SNE~\cite{van2008visualizing} to visualize the output token feature distributions at the first block, the tokens selected by ToE, and the output tokens after the second block.
    Baselines are DeiT-small trained on ImageNet-1K.
    ToE preserves the distribution integrity of intermediate features of the original token set across different Transformer blocks while ensuring that feature distributions are as wide as possible.
    }
    \label{fig:t_sne}
\end{figure*}

\subsubsection{Feature-distribution Token Merging}
\label{sssec:fdtm}
After token expansion, we aim to retain the effective information of the unselected tokens, such that we merge the unselected tokens that the feature distributions are close to the selected ones.
The feature-distribution token merging can be formulated as:
\begin{equation}
\small
    \begin{array}{c}
        \mathcal{S}_{l}^{'} =\{\text{mean}(\mathbb{A}_j,\mathcal{S}_{j}^{(M)})|\forall j\in\{1,2,\cdots,|\mathbb{A}|\}\}, \text{where}\\
        \mathcal{S}_{j}^{(M)} =\{\mathbb{B}_{i}| \mathcal{I}_{i}^{(M)}==j, \forall i\in\{1,2,\cdots,|\mathbb{B}|\}\},  \\
        \mathcal{I}^{(M)} =\text{argmin}_j(\mathcal{D}(\mathbb{B}, \mathbb{A})_{i,j}),                                    \\
    \end{array}
    \label{eq7}
\end{equation}
where $\mathcal{S}_{l}^{'} \in\mathbb{R}^{|\mathbb{A}|\times d} $ is the token set merging the closest tokens from $\mathbb{B}$ to $\mathbb{A}$, and $\text{mean}(\mathbb{A}_j,\mathcal{S}_{j}^{(M)})$ indicate that we merge  $\mathbb{B}$ into $\mathbb{A}$ \textit{averagely} based on the indice set $\mathcal{I}^{(M)} \in\mathbb{R}^{|\mathbb{B}|}$. Note that every $\mathbb{B}_i$ participates in the merging to avoid the information dropping for the unselected tokens.

\subsubsection{Optimization of ToE}

Our objective loss is the same as the original models, \eg, cross-entropy loss in DeiT.
The training details of \OurMethod are presented in Algorithm~\ref{algorithm1}. Note that we only apply \OurMethod to the output tokens of the first transformer block.
The detailed analysis is discussed in Sec.~\ref{subsec:ablation study}.

\OurMethod is a plug-and-play acceleration module, which has three following advantages: 
(1) As shown in Fig.~\ref{fig:t_sne}, %
we observed that the selected token set obtained by \OurMethod in the multiple block outputs has a larger average distribution distance via T-SNE~\cite{van2008visualizing}, compared to that in the original full-token set (see First block \emph{vs.} After ToE).
Moreover, it maintains a feature distribution similar to the original token set.
It indicates \OurMethod can preserve the integrity of the intermediate feature distribution of the original token set across different Transformer blocks by reducing the number of tokens.
(2) \OurMethod is a parameter-free module, it does not introduce any trainable parameters and utilizes efficient matrix calculations that the computational overhead is negligible, compared to computation-intensive self-attention.
(3) The speedup factors (\eg, token kept rate $r_1$ and training stage $N_g$) of \OurMethod are independent of the original model's training hyper-parameters.
This decoupling allows \OurMethod to be seamlessly integrated into the training process of the original model, obviating the need for any adjustments to the training hyper-parameters.

\begin{algorithm}[t]\footnotesize
    \caption{Optimization with ToE}
    \label{algorithm1}
    \KwIn{
        Input dataset $\mathcal{X}$, output token number $N_l$, total training stage $N_g$,
        kept rate of the first training stage $r_1$,
        repetition step of the parallel expanding operation $k$, Transformer parameters $\theta$, maximum iterations $T$.
    }
    \KwOut{Updated Transformer parameters $\theta$}
    \For{$t\leftarrow 1$ \KwTo $T$}{
        Sample from $\mathcal{X}$ to obtain data sample $x$, feed-forwarded through the embedding and first $l$-th transformer blocks to obtain the output token set $\mathcal{S}_l$\;

        \%\%\%\emph{Spatial-distribution Token Initialization}\%\%\%\\
        $r_0 \leftarrow  \frac{1}{2} r_1$; \\
        Initialize $\mathbb{A}$ and $\mathbb{B}$ by $r_0$, $\mathcal{S}_l$ via Eq.~\ref{eq2}\;

        \%\%\%\emph{Widest Feature-distribution Token Expansion}\%\%\%\\
        Obtain the current training stage $\delta=\lceil N_g*t/T\rceil$\;
        \For{$m\leftarrow 1$ \KwTo $\delta$}{
            \eIf{$m=1$}
            {$\mu_m\leftarrow r_1-r_0$\;}
            {$\mu_m\leftarrow \frac{1-r_1}{N_g-1}$}

            \For{$n\leftarrow 1$ \KwTo $k$}{
                Update $\mathbb{A}$ and $\mathbb{B}$ by $\mu_m$, $N_l$, $k$, prior $\mathbb{A}$ and prior $\mathbb{B}$ via Eq.~\ref{eq6}\;
            }
        }

        \%\%\%\emph{Feature-distribution Token Merging}\%\%\%\\
        Obtain $\mathcal{S}_{l}^{'}$ by $\mathbb{A}$ and $\mathbb{B}$ via Eq.~\ref{eq7}\;

        $\mathcal{S}_{l}^{'}$ feed-forwarded through the $l$+1-th transformer block to final layer and progressively obtain the final prediction $y$\;

        \%\%\%\emph{Parameter Updating}\%\%\%\\
        Use $y$ to compute the loss and obtain the gradient $\nabla\theta$\;
        Use $\nabla\theta$ to update prior $\theta$ via the optimizer to obtain new $\theta$\;
    }
    \Return{$\theta$}
\end{algorithm}

\section{Experiments}
\label{sec:experiments}

\subsection{Experimental Settings}
\label{subsec:experimental settings}

\textbf{Datasets and baselines.}
We evaluate our method on ImageNet-1K~\cite{deng2009imagenet} and CIFAR-10/100~\cite{krizhevsky2009learning}.
For baselines, we use two popular ViTs, \ie, DeiT~\cite{touvron2021training} and LV-ViT~\cite{jiang2021all}, as the base models to evaluate the proposed \OurMethod on ImageNet-1K.
To further evaluate the universality, we integrate \OurMethod into the efficient training framework EfficientTrain~\cite{wang2023efficienttrain}.
Moreover, we evaluate the transfer learning ability using pre-trained weights of \OurMethod on DeiT and the performance of accelerating the fine-tuning process with \OurMethod on CIFAR-10/100.

\textbf{Evaluation metrics.}
We report Top-1 accuracy, the GPU training time and FLOPs as the evaluation metric.
To evaluate the training speed, we report the total GPU hours consumed during the entire training process, as well as the theoretical FLOPs for \emph{one forward-backward process}.
To avoid the impact of memory access and kernel launching on training time~\cite{ding2023network}, we report the GPU hours on different numbers of GPUs,
but with the same GPU numbers to evaluate different training methods.
The FLOPs for the forward process are measured using thop~\footnote{https://github.com/Lyken17/pytorch-OpCounter/blob/master/thop}, and for the backward process, we follow~\cite{epoch2021backwardforwardFLOPratio} and calculate it as twice the FLOPs of the forward process.

\textbf{Implementations.}
All methods are trained by Pytorch~\cite{paszke2019pytorch}.
For DeiT and LV-ViT, all experiments are conducted on four NVIDIA RTX A6000 GPUs\footnote{Note that the used number of GPUs for training may be different to the evaluation of training speedup for a fair comparison.}, while EfficientTrain is trained on eight NVIDIA RTX A6000 GPUs.
All hyper-parameters (\eg, learning rate, decay strategy and rate), and training strategies and optimization processes are the same as the original papers unless otherwise specified. 

\textbf{Growth strategy.}
In default, we divide the origin training process into $N_g=3$ stages on average.
The token kept rate of 1st stage $r_1$ is set to $0.4$, $0.5$ or $0.6$, our method is corresponding to be denoted as \OurMethod$_{r_1=0.4}$, \OurMethod$_{r_1=0.5}$ or \OurMethod$_{r_1=0.6}$. Correspondingly, the kept rate of the initial stage $r_0$ is set to 0.2, 0.25 and 0.3. 
The repetition step of parallel expanding operation $k$ is default set to $2$, and we perform \OurMethod on the output tokens of the first block for all models.

\subsection{Results on ImageNet-1k}
\label{subsec:results on imageNet-1k}

\begin{table*}[htbp]
    \caption{Performance comparison for DeiT on ImageNet-1K. a/b in the column of Top-1 Acc. means without/with EMA strategy using the official GitHub repo$^{\dag}$.
        The training time is averagely measured on one/two/four NVIDIA RTX A6000 GPUs for DeiT-tiny/small/base 3 times, and the batch size is set to $1,024$ in all following tables and figures.}
    \centering
    \resizebox{0.9\linewidth}{!}{
        \begin{threeparttable}
            \begin{tabular}{c|c|ccc|cccc}
                \toprule
                \multirow{2}{*}{Model}      & \multirow{2}{*}{Method}                                                   & \multicolumn{3}{c|}{Consistency}  & \multirow{2}{*}{Top-1 Acc. (\%)}  & \multicolumn{1}{c}{GFLOPs}        & \multicolumn{1}{c}{Training time}         & \multicolumn{1}{c}{Acceleration}                                                                                      \\
                \cmidrule{3-5}
                                            &                                                                           & Hyper?                            & Architecture?                     & Strategy?                         &                                           & (per training iter)                                 & (total GPU hours)           & (practical rate)                  \\
                \midrule
                \multirow{5}{*}{DeiT-tiny}  & Baseline~\cite{touvron2021training}                                       & $-$                               & $-$                               & $-$                               & 72.2                                      & $3.3\times10^3$                                     & 54.6h                       & 1.00$\times$                      \\
                                            & (NeurIPS'21) S$^2$ViTE-Tiny (600 epoch)~\cite{chen2021chasing}            & $\times$                          & $\times$                          & $\checkmark$                      & 70.1 (-2.1)                               & $2.5\times10^3$ (1.32$\times$)                      & $-$                         & 1.19$\times$                  \\
                                            & (ICLR'23) ToMe$_{r_{8}\rightarrow}^{\mathrm{DeiT}}$~\cite{bolya2022token} & $\checkmark$                      & $\times$                          & $\checkmark$                      & 71.7 (-0.5)                               & $2.5\times10^3$ (1.32$\times$)                      & 53.3h                       & 1.02$\times$                      \\
                                            & (CVPR'23) NetworkExpansion$_{6\rightarrow 12}$~\cite{ding2023network}     & $\checkmark$                      & $\checkmark$                      & $\times$                          & 70.3 (-1.9) / 70.1 (-2.1)                 & $2.5\times10^3$ (1.32$\times$)                      & 43.2h                       & 1.26$\times$                      \\
                                            & \cellcolor[gray]{0.8}\OurMethod$_{r_1=0.5}$ (Ours)                        & \cellcolor[gray]{0.8}$\checkmark$ & \cellcolor[gray]{0.8}$\checkmark$ & \cellcolor[gray]{0.8}$\checkmark$ & \cellcolor[gray]{0.8}\textbf{72.6 (+0.4)} & \cellcolor[gray]{0.8}$2.6\times10^3$ (1.27$\times$) & \cellcolor[gray]{0.8}44.2h  & \cellcolor[gray]{0.8}1.24$\times$ \\
                \midrule
                \multirow{4}{*}{DeiT-small} & Baseline~\cite{touvron2021training}                                       & $-$                               & $-$                               & $-$                               & 79.8                                      & $1.3\times10^4$                      & 124.5h                      & 1.00$\times$                      \\
                                            & (ICLR'23) ToMe$_{r_{8}\rightarrow}^{\mathrm{DeiT}}$~\cite{bolya2022token} & $\checkmark$                      & $\times$                          & $\checkmark$                      & 79.7 (-0.1)                               & $9.8\times10^3$ (1.33$\times$)                      & 121.5h                      & 1.02$\times$                      \\
                                            & (CVPR'23) NetworkExpansion$_{6\rightarrow 12}$~\cite{ding2023network}     & $\checkmark$                      & $\checkmark$                      & $\times$                          & 78.8 (-1.0) / 78.6 (-1.2)                 & $9.8\times10^3$ (1.33$\times$)                      & 100.3h                      & 1.24$\times$                      \\
                                            & \cellcolor[gray]{0.8}\OurMethod$_{r_1=0.5}$ (Ours)                        & \cellcolor[gray]{0.8}$\checkmark$ & \cellcolor[gray]{0.8}$\checkmark$ & \cellcolor[gray]{0.8}$\checkmark$ & \cellcolor[gray]{0.8}\textbf{79.8 (+0.0)} & \cellcolor[gray]{0.8}$1.0\times10^4$ (1.30$\times$) & \cellcolor[gray]{0.8}102.2h & \cellcolor[gray]{0.8}1.22$\times$ \\

                \midrule
                \multirow{7}{*}{DeiT-base}  & Baseline~\cite{touvron2021training}                                       & $-$                               & $-$                               & $-$                               & 81.8                                      & $5.2\times10^4$                      & 292.8h                      & 1.00$\times$                      \\
                                            & (ICML'19) StackBERT~\cite{gong2019efficient}                              & $\checkmark$                      & $\checkmark$                      & $\times$                          & 80.8 (-1.0)                               & $4.2\times10^4$ (1.24$\times$)                      & 231.6h                      & 1.26$\times$                      \\
                                            & (CVPR'23) NetworkExpansion$_{6\rightarrow 12}$~\cite{ding2023network}     & $\checkmark$                      & $\checkmark$                      & $\times$                          & 81.0 (-0.8) / 81.5 (-0.3)                 & $3.9\times10^4$ (1.33$\times$)                      & 226.8h                      & 1.29$\times$                      \\
                                            & \cellcolor[gray]{0.8}\OurMethod$_{r_1=0.5}$ (Ours)                        & \cellcolor[gray]{0.8}$\checkmark$ & \cellcolor[gray]{0.8}$\checkmark$ & \cellcolor[gray]{0.8}$\checkmark$ & \cellcolor[gray]{0.8}\textbf{81.6 (-0.2)} & \cellcolor[gray]{0.8}$4.0\times10^4$ (1.30$\times$) & \cellcolor[gray]{0.8}231.2h & \cellcolor[gray]{0.8}1.27$\times$ \\
                                            & \cellcolor[gray]{0.8}\OurMethod$_{r_1=0.4}$ (Ours)                        & \cellcolor[gray]{0.8}$\checkmark$ & \cellcolor[gray]{0.8}$\checkmark$ & \cellcolor[gray]{0.8}$\checkmark$ & \cellcolor[gray]{0.8}81.4 (-0.4)          & \cellcolor[gray]{0.8}$3.8\times10^4$ (1.37$\times$) & \cellcolor[gray]{0.8}225.2h & \cellcolor[gray]{0.8}1.30$\times$ \\
                                            & \cellcolor[gray]{0.8}\OurMethod$_{r_1=0.5}^{\mathrm{Hyper}}$ (Ours)       & \cellcolor[gray]{0.8}$\times$     & \cellcolor[gray]{0.8}$\checkmark$ & \cellcolor[gray]{0.8}$\checkmark$ & \cellcolor[gray]{0.8}\textbf{81.8 (+0.0)} & \cellcolor[gray]{0.8}$3.6\times10^4$ (1.44$\times$) & \cellcolor[gray]{0.8}213.2h & \cellcolor[gray]{0.8}1.37$\times$ \\
                                            & \cellcolor[gray]{0.8}\OurMethod$_{r_1=0.4}^{\mathrm{Hyper}}$ (Ours)       & \cellcolor[gray]{0.8}$\times$     & \cellcolor[gray]{0.8}$\checkmark$ & \cellcolor[gray]{0.8}$\checkmark$ & \cellcolor[gray]{0.8}81.7 (-0.1)          & \cellcolor[gray]{0.8}$3.3\times10^4$ (1.58$\times$) & \cellcolor[gray]{0.8}202.8h & \cellcolor[gray]{0.8}1.44$\times$ \\
                \bottomrule
            \end{tabular}
            \begin{tablenotes}
                \item[$\dag$] https://github.com/huawei-noah/Efficient-Computing/tree/master/TrainingAcceleration/NetworkExpansion
            \end{tablenotes}
        \end{threeparttable}
    }
    \label{tab:deit_imagenet}
\end{table*}

\begin{table}[htbp]
    \caption{Performance comparison for LV-ViT on ImageNet-1K.
        $\ddag$ indicates that results reproduced by the official GitHub repo.
        The training time is averagely measured on two/four NVIDIA RTX A6000 GPUs 3 times for LV-ViT-T/S with a fixed batch size of $1,024$.
    }
    \centering
    \resizebox{0.99\linewidth}{!}{
        \begin{tabular}{c|c|ccc}
            \toprule
            \multirow{2}{*}{Model}    & \multirow{2}{*}{Method}                                                & \multirow{2}{*}{Top-1 Acc. (\%)}          & \multicolumn{1}{c}{GFLOPs}                          & \multicolumn{1}{c}{Training time}          \\
                                      &                                                                        &                                           & (per training iter)                                 & (total GPU hours)                          \\
            \midrule
            \multirow{3}{*}{LV-ViT-T} & Baseline~\cite{jiang2021all}                                           & 79.1                                      & $8.2\times10^3$                                     & 130.5h                                     \\
                                      & (CVPR'23) NetworkExpansion$_{6\rightarrow 12}$~\cite{ding2023network}  & ${\ddag}$78.8 (-0.3)                      & $7.1\times10^3$ (1.15$\times$)                      & 114.4h (1.14$\times$)                      \\
                                      & \cellcolor[gray]{0.8}ToE$_{r_1=0.4}$ (Ours)                            & \cellcolor[gray]{0.8}\textbf{79.4 (+0.3)} & \cellcolor[gray]{0.8}$7.0\times10^3$ (1.17$\times$) & \cellcolor[gray]{0.8}113.9h (1.15$\times$) \\
            \midrule
            \multirow{3}{*}{LV-ViT-S} & Baseline~\cite{jiang2021all}                                           & 83.3                                      & $1.9\times10^4$                                     & 237.3h                                     \\
                                      & (CVPR'23) NetworkExpansion$_{8\rightarrow 16}$~\cite{ding2023network}  & ${\ddag}$82.9 (-0.4)                      & $1.5\times10^4$ (1.27$\times$)                      & 195.5h (1.21$\times$)                      \\
                                      & \cellcolor[gray]{0.8}ToE$_{r_1=0.4}$ (Ours)                            & \cellcolor[gray]{0.8}\textbf{83.3 (+0.0)} & \cellcolor[gray]{0.8}$1.4\times10^4$ (1.36$\times$) & \cellcolor[gray]{0.8}195.3h (1.22$\times$) \\
            \midrule
            \multirow{3}{*}{LV-ViT-M} & Baseline~\cite{jiang2021all}                                           & 84.1                                      & $3.7\times10^4$                                     & 368.7h                                     \\
                                      & (CVPR'23) NetworkExpansion$_{10\rightarrow 20}$~\cite{ding2023network} & 84.0 (-0.1)                               & $2.9\times10^4$ (1.28$\times$)                      & 292.7h (1.26$\times$)                      \\
                                      & \cellcolor[gray]{0.8}\OurMethod$_{r_1=0.4}$ (Ours)                     & \cellcolor[gray]{0.8}\textbf{84.1 (+0.0)} & \cellcolor[gray]{0.8}$2.7\times10^4$ (1.37$\times$) & \cellcolor[gray]{0.8}292.5h (1.26$\times$) \\

            \bottomrule
        \end{tabular}

    }
    \label{tab:lvvit_imagenet}
\end{table}

\textbf{DeiT and LV-ViT}
As shown in Tab.~\ref{tab:deit_imagenet}, \OurMethod achieves lossless training acceleration with SOTA performance.
For example, \OurMethod$_{r_1=0.5}$ achieves $0.4\%$ Top-1 accuracy improvement with $1.27\times$ theoretical and $1.24\times$ practical faster speed to train DeiT-tiny. For DeiT-small, it achieves $1.3\times$ training acceleration without accuracy drop.
Compared to the SOTA methods, \OurMethod$_{r_1=0.5}$ outperforms SViTE~\cite{chen2021chasing} and NetworkExpansion~\cite{ding2023network} at least $1\%$ Top-1 accuracy at the consistent acceleration ratio for training both DeiT-tiny and  DeiT-small.
Compared to ToMe~\cite{bolya2022token}, \OurMethod$_{r_1=0.5}$ also achieves both higher accuracy and practical training speed. Note that ToMe is able to reduce GFLOPs, but fails to accelerate training due to the usage of unfriendly weighted average attention and layer-wise merging operations.
For DeiT-base, \OurMethod$_{r_1=0.5}$ drops only $0.2\%$ Top-1 accuracy while saving more than $60$ GPU hours in the practical training process, which is comparable to NetworkExpansion with EMA.
If we relax the restriction of hyper-parameter consistency (presented in Appendix), \OurMethod$_{r_1=0.4}^{\mathrm{Hyper}}$ outperforms NetworkExpansion with 0.2\% accuracy and 24h training time reduction.

For LV-ViT-T and LV-ViT-S shown in Tab.~\ref{tab:lvvit_imagenet}, \OurMethod$_{r_1=0.4}$ achieves efficient training with 1.2$\times$ acceleration rate, while without accuracy drop or even with accuracy improvement for training LV-ViT-T, compared to baselines. 
Note that the results of \OurMethod$_{r_1=0.4}$ and NetworkExpansion are reported with EMA, due to the default LV-ViT training with EMA.
In addition, \OurMethod$_{r_1=0.4}$ outperforms NetworkExpansion in both training acceleration and accuracy with 0.5h training time reduction and 0.6\% accuracy for LV-ViT-T, respectively.

\begin{figure}[t]
    \centering
    \includegraphics[scale = 0.3]{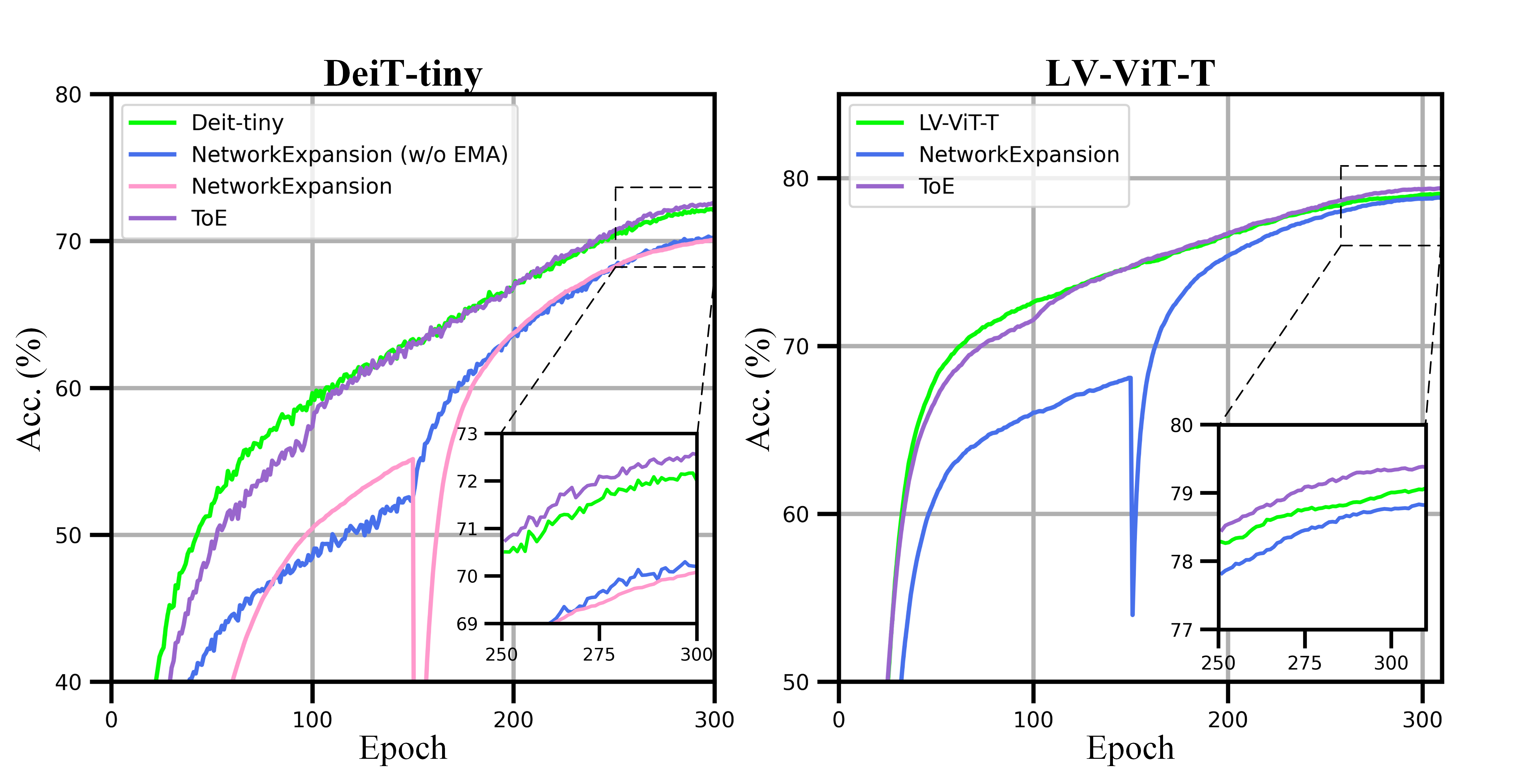}
    \caption{Validation Top-1 accuracy of DeiT-tiny and LV-ViT-T on ImageNet-1k during training with different methods.} %
    \label{fig:deit_and_lvvit_acc}
\end{figure}

We also present the validation Top-1 accuracy of \OurMethod and NetworkExpansion during training DeiT-tiny and LV-ViT-T in Fig.~\ref{fig:deit_and_lvvit_acc}.
As observed, \OurMethod initially reduces token redundancy during training, resulting in some performance drops compared to the baseline.
However, in the later stages of training, \OurMethod introduces more tokens for training, gradually reducing the accuracy gap to the baseline.
Benefiting from the reduction of token redundancy in the early stages, models trained by \OurMethod with the proposed token expansion and merging achieve higher accuracies, compared to baselines. Compared to NetworkExpansion, our \OurMethod is more stable to train with consistent accuracy improvement during training, while the accuracy of NetworkExpansion with EMA drops significantly at the intermediate epoch number and then restores due to the inconsistent structures of before-and-after models when structure growing.  
More validation curves are presented in the Appendix.

\begin{table}[t]
    \caption{Performance comparison between EfficientTrain~\cite{wang2023efficienttrain} and our combination framework on ImageNet-1K.} %
    \centering
    \resizebox{0.99\linewidth}{!}{
        \begin{tabular}{c|c|ccc}
            \toprule
            \multirow{2}{*}{Model}      & \multirow{2}{*}{Method}                                             & \multirow{2}{*}{Top-1 Acc. (\%)}          & GFLOPs                                              & Training time                             \\
                                        &                                                                     &                                           & (per training iter)                                 & (total GPU hours)                         \\
            \midrule
            \multirow{3}{*}{DeiT-tiny}  & Baseline (EfficientTrain)~\cite{wang2023efficienttrain}             & 72.5                                      & $1.3\times10^4$                                     & 52.5h                                     \\
                                        & (ICCV'23) EfficientTrain~\cite{wang2023efficienttrain}              & 73.3 (+0.8)                               & $8.8\times10^3$ (1.48$\times$)                      & 36.5h (1.44$\times$)                      \\
                                        & \cellcolor[gray]{0.8}EfficientTrain + \OurMethod$_{r_1=0.6}$ (Ours) & \cellcolor[gray]{0.8}\textbf{73.5 (+1.0)} & \cellcolor[gray]{0.8}$7.6\times10^3$ (1.71$\times$) & \cellcolor[gray]{0.8}32.3h (1.63$\times$) \\
            \midrule
            \multirow{3}{*}{DeiT-small} & Baseline (EfficientTrain)~\cite{wang2023efficienttrain}             & 80.3                                      & $5.2\times10^4$                                     & 121.3h                                    \\
                                        & (ICCV'23) EfficientTrain~\cite{wang2023efficienttrain}              & \textbf{80.4 (+0.1)}                      & $3.4\times10^4$ (1.53$\times$)                      & 85.2h (1.42$\times$)                      \\
                                        & \cellcolor[gray]{0.8}EfficientTrain + \OurMethod$_{r_1=0.6}$ (Ours) & \cellcolor[gray]{0.8}\textbf{80.4 (+0.1)} & \cellcolor[gray]{0.8}$2.9\times10^4$ (1.79$\times$) & \cellcolor[gray]{0.8}79.4h (1.53$\times$) \\

            \bottomrule
        \end{tabular}

    }
    \label{tab:et_train_imagenet}
\end{table}

\nbf{Combination with EfficientTrain~\cite{wang2023efficienttrain}}
\OurMethod can be seamlessly integrated into the EfficientTrain framework to further improve the performance.
We do not modify the pipeline of EfficientTrain and simply apply \OurMethod to the output tokens of the model's first block.
The results are summarized in Tab.~\ref{tab:et_train_imagenet}, which effectively evaluates the universality of \OurMethod.
The combination of EfficientTrain and \OurMethod achieves higher training speeds to further enhance the training efficiency of EfficientTrain with accuracy improvement.

\subsection{Transfer Results on CIFAR-10/100}
we further explore the transfer learning ability of \OurMethod-pre-trained weights and evaluate whether \OurMethod can be used to accelerate the fine-tuning on CIFAR-10/100.
For the fine-tuning settings, we follow the settings of the official GitHub repo~\footnote{https://github.com/facebookresearch/deit}.
We introduce the training details in the Appendix.

As shown in Tab.~\ref{tab:finetune}, pre-training weights by \OurMethod is able to improve the accuracy on CIFAR-10/100 for DeiT-tiny/small, when using the same baseline training for fine-tuning (see the 1st and 3rd rows in both DeiT-tiny and DeiT-small). For example, ToE pre-training outperforms baseline pre-training by 0.32\% accuracy on CIFAR-100, which evaluates the strong transfer ability of \OurMethod. In addition, our \OurMethod is also effective and efficient for fine-tuning (see the 1st and 2nd rows in DeiT-tiny/small). \OurMethod achieves 1.3$\times$ acceleration for fine-tuning DeiT-tiny with 0.03 accuracy improvement on CIFAR-10. Further, we employ \OurMethod for both pre-training and fine-tuning, which significantly accelerates the training with an accuracy improvement of at least 0.06\% on CIFAR-10 for both DeiT-tiny/small, compared to that using both baselines.

\begin{table}[t]
    \caption{Results for fine-tuning DeiT on CIFAR-10/100.}
    \centering
    \resizebox{0.99\linewidth}{!}{
        \begin{tabular}{c|cc|cc|cc}
            \toprule
            \multirow{2}{*}{Model}      & \multicolumn{2}{c|}{Pre-training}                        & \multicolumn{2}{c|}{Fine-tuning} & \multicolumn{2}{c}{Top-1 Acc. (\%)}                                                                                                                                              \\
            \cmidrule{2-7}
                                        & Method                                                   & Acceleration                     & Method                                                   & Acceleration              & CIFAR-10                                    & CIFAR-100                                   \\
            \midrule
            \multirow{4}{*}{DeiT-tiny}  & Baseline~\cite{touvron2021training}                      & 1.0x                             & Baseline~\cite{touvron2021training}                      & 1.0x                      & 98.07                                       & 86.78                                       \\
                                        & \cellcolor[gray]{0.8}Baseline~\cite{touvron2021training} & \cellcolor[gray]{0.8}1.0x        & \cellcolor[gray]{0.8}ToE$_{r_1=0.5}$                     & \cellcolor[gray]{0.8}1.3x & \cellcolor[gray]{0.8}98.10 (+0.03)          & \cellcolor[gray]{0.8}86.74 (-0.04)          \\
                                        & \cellcolor[gray]{0.8}ToE$_{r_1=0.5}$                     & \cellcolor[gray]{0.8}1.3x        & \cellcolor[gray]{0.8}Baseline~\cite{touvron2021training} & \cellcolor[gray]{0.8}1.0x & \cellcolor[gray]{0.8}\textbf{98.19 (+0.12)} & \cellcolor[gray]{0.8}\textbf{87.10 (+0.32)} \\
                                        & \cellcolor[gray]{0.8}ToE$_{r_1=0.5}$                     & \cellcolor[gray]{0.8}1.3x        & \cellcolor[gray]{0.8}ToE$_{r_1=0.5}$                     & \cellcolor[gray]{0.8}1.3x & \cellcolor[gray]{0.8}98.16 (+0.09)          & \cellcolor[gray]{0.8}86.91 (+0.13)          \\
            \midrule
            \multirow{4}{*}{DeiT-small} & Baseline~\cite{touvron2021training}                      & 1.0x                             & Baseline~\cite{touvron2021training}                      & 1.0x                      & 98.93                                       & 90.15                                       \\
                                        & \cellcolor[gray]{0.8}Baseline~\cite{touvron2021training} & \cellcolor[gray]{0.8}1.3x        & \cellcolor[gray]{0.8}ToE$_{r_1=0.5}$                     & \cellcolor[gray]{0.8}1.3x & \cellcolor[gray]{0.8}98.96 (+0.03)          & \cellcolor[gray]{0.8}90.19 (+0.04)          \\
                                        & \cellcolor[gray]{0.8}ToE$_{r_1=0.5}$                     & \cellcolor[gray]{0.8}1.3x        & \cellcolor[gray]{0.8}Baseline~\cite{touvron2021training} & \cellcolor[gray]{0.8}1.0x & \cellcolor[gray]{0.8}\textbf{99.03 (+0.10)} & \cellcolor[gray]{0.8}\textbf{90.37 (+0.22)} \\
                                        & \cellcolor[gray]{0.8}ToE$_{r_1=0.5}$                     & \cellcolor[gray]{0.8}1.3x        & \cellcolor[gray]{0.8}ToE$_{r_1=0.5}$                     & \cellcolor[gray]{0.8}1.3x & \cellcolor[gray]{0.8}98.99 (+0.06)          & \cellcolor[gray]{0.8}90.26 (+0.11)          \\

            \bottomrule
        \end{tabular}
    }
    \label{tab:finetune}
\end{table}

\subsection{Ablation Study}
\label{subsec:ablation study}

\textbf{Effect of speedup factors in \OurMethod.}
As presented in Tab.~\ref{tab:hyper}, we verify the sensitivity of the speedup factors mentioned in Sec.~\ref{subsec:token expansion}, such as the ratio of $r_0/r_1$, training stages $N_g$ and parallel expanding operation $k$.
At almost the same training time, \OurMethod is relatively insensitive to these factors, \emph{w.r.t} accuracy.
It allows \OurMethod to be easily integrated into the different models' training pipeline with minimal factor adjustments.

\begin{figure}[t]
    \centering
    \includegraphics[scale = 0.65]{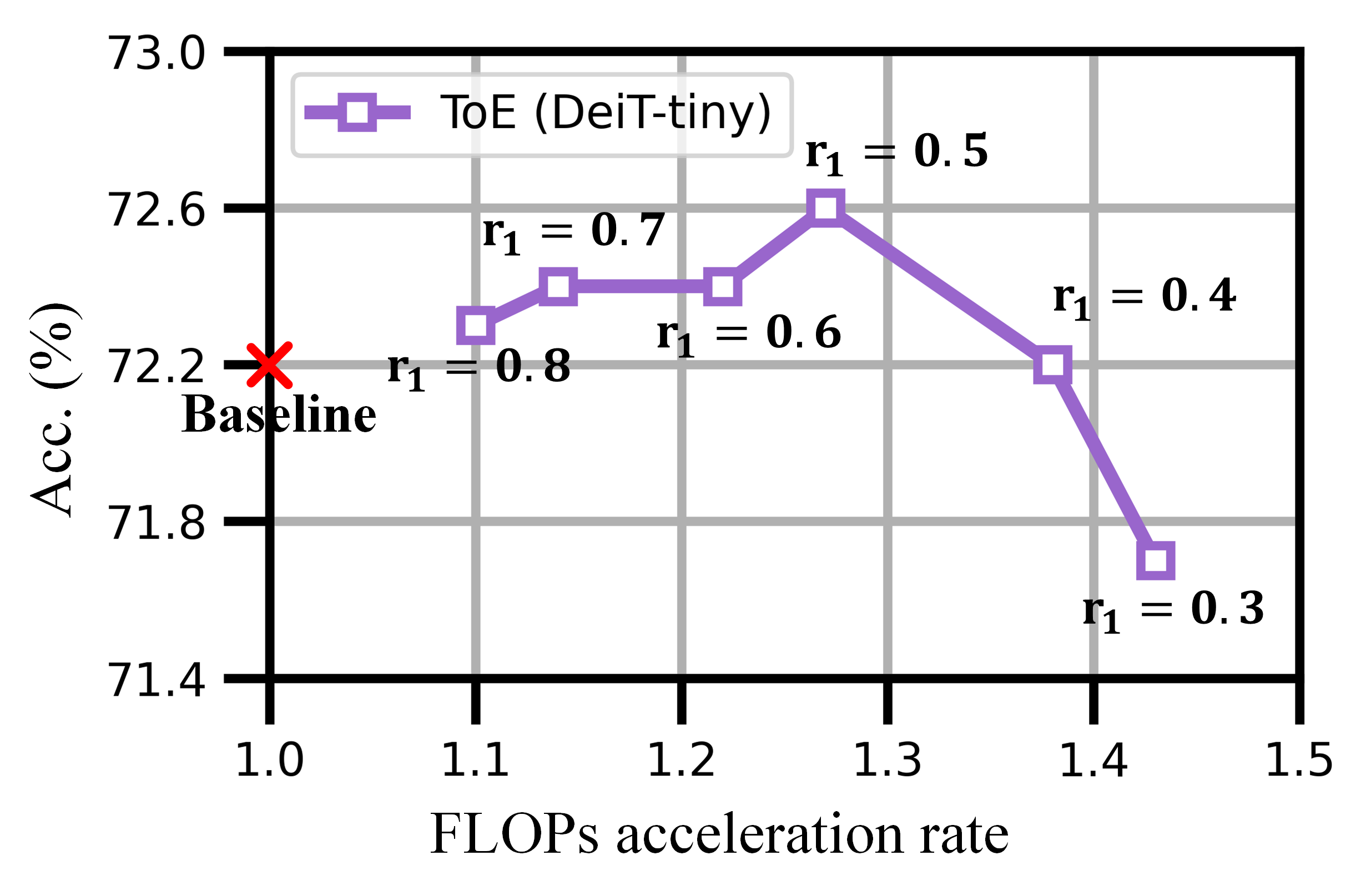}
    \caption{Trade-off between acceleration ratio and model performance by setting different $r_1$.}
    \label{fig:progressive_acc}
\end{figure}

We further adjust the keep rate of the first stage $r_1$ to control the training speed, and the relationship between $r_1$ and training speed is illustrated in Fig.~\ref{fig:progressive_acc}.
We found \OurMethod achieves more than $1.3\times$ acceleration on DeiT-tiny without accuracy dropping.
Additionally, it also demonstrates that reducing token redundancy in the early stages of training sometimes improves the model performance.

\begin{table}[t]
    \caption{Ablation studies of different speedup factors for DeiT-tiny on ImageNet-1K.
        The default $r_0/r_1$, $N_g$ and $k$ are set to $1/2$, $3$ and $2$, respectively.
        All results in this table have almost the same training speeds for 44h training (total GPU hours).
    }
    \centering
    \resizebox{0.99\linewidth}{!}{
        \begin{tabular}{c|c|cccccc>{\columncolor[gray]{0.8}}c}
            \toprule
            \multirow{3}{*}{DeiT-tiny} & Factors           & $r_0/r_1=1/3$ & $r_0/r_1=2/3$ & $N_g=2$ & $N_g=4$ & $\rm{k}=1$ & $\rm{k}=3$    & default       \\
            \cmidrule{2-9}
                                       & Top-1 Acc. (\%) & 72.3          & 72.5          & 72.4    & 72.5    & 72.5       & \textbf{72.6} & \textbf{72.6} \\
            \bottomrule
        \end{tabular}
    }
    \label{tab:hyper}
\end{table}

\begin{table}[t]
    \caption{Effect of ``initialization-expansion-merge'' pipeline  for DeiT on ImageNet-1K.
    $\pm$ indicates we conduct 3 runs to calculate the mean and std.
    }
    \centering
    \resizebox{0.7\linewidth}{!}{
        \begin{tabular}{cc|c|c|cc}
            \toprule
            \multicolumn{2}{c|}{Initialization} & \multirow{2}{*}{Expansion}        & \multirow{2}{*}{Merge}            & \multicolumn{2}{c}{Top-1 Acc. (\%)}                                                                           \\
            \cmidrule{1-2}\cmidrule{5-6}
            \multirow{1}{*}{Random}             & \multirow{1}{*}{Spatial}          &                                   &                                     & DeiT-tiny                          & DeiT-small                         \\
            \midrule
            \cellcolor[gray]{0.8}$\times$       & \cellcolor[gray]{0.8}$\checkmark$ & \cellcolor[gray]{0.8}$\checkmark$ & \cellcolor[gray]{0.8}$\checkmark$   & \cellcolor[gray]{0.8}\textbf{72.6} & \cellcolor[gray]{0.8}\textbf{79.8} \\
            $\checkmark$                        & $\times$                          & $\checkmark$                      & $\checkmark$                        & 72.3$\pm$0.2                       & 79.7$\pm$0.1                       \\
            $\times$                            & $\checkmark$                      & $\times$                          & $\checkmark$                        & 71.2                               & 79.1                               \\
            $\times$                            & $\checkmark$                      & $\checkmark$                      & $\times$                            & 71.7                               & 79.6                               \\
            \bottomrule
        \end{tabular}
    }
    \label{tab:strategy}
\end{table}

\textbf{Effect of ``Initialization-expansion-merging''.}
Tab.~\ref{tab:strategy} provides an analysis of the necessity of each step in the proposed ``initialization-expansion-merging'' pipeline.
When we randomly select tokens as the initial token set rather than \textit{spatial-distribution token initialization}, it leads to the performance degradation.
Furthermore, removing \textit{widest feature-distribution token expansion} and \textit{feature-distribution token merging} from the pipeline significantly decreases the accuracy, \eg, more than 0.9\% and 1.4\% accuracy drops without the merging and expansion for DeiT-tiny, respectively.

\textbf{Where to apply \OurMethod.}
Work in~\cite{raghu2021vision, pan2022budgeted} demonstrates that class attention tends to be a global pooling as more attention operations are performed, and tokens in early blocks are more similar.
This leads to more redundancy in tokens from early blocks.
Consequently, applying \OurMethod to the output tokens of early blocks can achieve higher acceleration.
As shown in Tab.~\ref{tab:block_number}, we default apply \OurMethod to the output tokens of the first block, which achieves the best trade-off between accuracy and training speed, compared to other early blocks.

\begin{table}[t]
    \caption{Results of applying \OurMethod to different early transformer block's output tokens for DeiT-tiny on ImageNet-1K.
    }
    \centering
    \resizebox{0.7\linewidth}{!}{
        \begin{tabular}{c|ccc}
            \toprule
            \multirow{2}{*}{Block}           & Top-1 Acc. (\%)           & GFLOPs                                & Training time              \\
                                             & DeiT-tiny                 & (per training iter)                   & (total GPU hours)          \\
            \midrule
            Embedding                        & 72.1                      & $2.51\times10^3$                      & 43.5h                      \\
            \cellcolor[gray]{0.8}First block & \cellcolor[gray]{0.8}\textbf{72.6} & \cellcolor[gray]{0.8}$2.58\times10^3$ & \cellcolor[gray]{0.8}44.2h \\
            Second block                     & 72.2                      & $2.65\times10^3$                      & 45.2h                      \\
            Third block                      & 72.1                      & $2.71\times10^3$                      & 46.9h                      \\
            \bottomrule
        \end{tabular}
    }
    \label{tab:block_number}
\end{table}

\begin{table}[t]
    \caption{Results of different feature-distribution distances in Eq.~\ref{eq3} for DeiT on ImageNet-1K.}
    \centering
    \resizebox{0.55\linewidth}{!}{
        \begin{tabular}{c|cc}
            \toprule
            \multirow{2}{*}{Measure}             & \multicolumn{2}{c}{Top-1 Acc. (\%)}                                      \\
            \cmidrule{2-3}
                                                 & DeiT-tiny                           & DeiT-small                         \\
            \midrule
            Manhattan Distance                   & 69.8                                & 78.0                               \\
            Euclidean Distance                   & 70.6                                & 78.4                               \\
            \cellcolor[gray]{0.8}Cosine Distance & \cellcolor[gray]{0.8}\textbf{72.6}  & \cellcolor[gray]{0.8}\textbf{79.8} \\
            \bottomrule
        \end{tabular}
    }
    \label{tab:distance}
\end{table}

\textbf{Effect of the feature-distribution distance.}
We explore the metric that measures the feature-distribution distance between two tokens in Eq.~\ref{eq3}.
As shown in Tab.~\ref{tab:distance}, we use three different metrics: Manhattan distance, Euclidean distance, and Cosine distance.
We observe that Cosine distance achieves the best performance as the distance metric.

\section{Conclusion}
\label{sec:conclusion}

In this paper, we proposed a novel token growth scheme \OurMethodFullName(\OurMethod) to achieve consistent training acceleration for ViTs. 
\OurMethod introduce an ``initialization-expansion-merging'' pipeline to maintain the integrity of the intermediate feature distribution of original transformers, preventing the loss of crucial learnable information in the training process. In experiments,
\OurMethod can be seamlessly integrated into the training of various transformers and efficient training frameworks in a lossless manner or even accuracy improvement, compared to the entire full-token training. These experimental results of \OurMethod also demonstrate the superior performance gains over the SOTA methods.

\section*{Acknowledgements}
This work is supported by the National Natural Science Foundation of China (NO. 62102151), the National Key Research and Development Program of China (No. 2023YFC3306401), Shanghai Sailing Program (21YF1411200), Shanghai Science and Technology Commission (22511104600), CCF-Tencent Rhino-Bird Open Research Fund, the Open Research Fund of Key Laboratory of Advanced Theory and Application in Statistics and Data Science, Ministry of Education (KLATASDS2305), the Fundamental Research Funds for the Central Universities. 

{\small
    \bibliographystyle{unsrt}
    \bibliography{11_references}

\begin{thebibliography}{10}

\bibitem{vaswani2017attention}
Ashish Vaswani, Noam Shazeer, Niki Parmar, Jakob Uszkoreit, Llion Jones, Aidan~N Gomez, {\L}ukasz Kaiser, and Illia Polosukhin.
\newblock Attention is all you need.
\newblock {\em NeurIPS}, 30, 2017.

\bibitem{kenton2019bert}
Jacob Devlin Ming-Wei~Chang Kenton and Lee~Kristina Toutanova.
\newblock Bert: Pre-training of deep bidirectional transformers for language understanding.
\newblock In {\em NAACL}, pages 4171--4186, 2019.

\bibitem{brown2020language}
Tom Brown, Benjamin Mann, Nick Ryder, Melanie Subbiah, Jared~D Kaplan, Prafulla Dhariwal, Arvind Neelakantan, Pranav Shyam, Girish Sastry, Amanda Askell, et~al.
\newblock Language models are few-shot learners.
\newblock {\em NeurIPS}, 33:1877--1901, 2020.

\bibitem{touvron2021training}
Hugo Touvron, Matthieu Cord, Matthijs Douze, Francisco Massa, Alexandre Sablayrolles, and Herv{\'e} J{\'e}gou.
\newblock Training data-efficient image transformers \& distillation through attention.
\newblock In {\em ICLR}, pages 10347--10357. PMLR, 2021.

\bibitem{jiang2021all}
Zi-Hang Jiang, Qibin Hou, Li~Yuan, Daquan Zhou, Yujun Shi, Xiaojie Jin, Anran Wang, and Jiashi Feng.
\newblock All tokens matter: Token labeling for training better vision transformers.
\newblock {\em NeurIPS}, 34:18590--18602, 2021.

\bibitem{carion2020end}
Nicolas Carion, Francisco Massa, Gabriel Synnaeve, Nicolas Usunier, Alexander Kirillov, and Sergey Zagoruyko.
\newblock End-to-end object detection with transformers.
\newblock In {\em ECCV}, pages 213--229. Springer, 2020.

\bibitem{xie2021segformer}
Enze Xie, Wenhai Wang, Zhiding Yu, Anima Anandkumar, Jose~M Alvarez, and Ping Luo.
\newblock Segformer: Simple and efficient design for semantic segmentation with transformers.
\newblock {\em NeurIPS}, 34:12077--12090, 2021.

\bibitem{dosovitskiy2020image}
Alexey Dosovitskiy, Lucas Beyer, Alexander Kolesnikov, Dirk Weissenborn, Xiaohua Zhai, Thomas Unterthiner, Mostafa Dehghani, Matthias Minderer, Georg Heigold, Sylvain Gelly, et~al.
\newblock An image is worth 16x16 words: Transformers for image recognition at scale.
\newblock In {\em ICLR}, 2020.

\bibitem{he2016deep}
Kaiming He, Xiangyu Zhang, Shaoqing Ren, and Jian Sun.
\newblock Deep residual learning for image recognition.
\newblock In {\em CVPR}, pages 770--778, 2016.

\bibitem{chen2021chasing}
Tianlong Chen, Yu~Cheng, Zhe Gan, Lu~Yuan, Lei Zhang, and Zhangyang Wang.
\newblock Chasing sparsity in vision transformers: An end-to-end exploration.
\newblock {\em NeurIPS}, 34:19974--19988, 2021.

\bibitem{bolya2022token}
Daniel Bolya, Cheng-Yang Fu, Xiaoliang Dai, Peizhao Zhang, Christoph Feichtenhofer, and Judy Hoffman.
\newblock Token merging: Your vit but faster.
\newblock In {\em ICLR}, 2022.

\bibitem{ding2023network}
Ning Ding, Yehui Tang, Kai Han, Chao Xu, and Yunhe Wang.
\newblock Network expansion for practical training acceleration.
\newblock In {\em CVPR}, pages 20269--20279, 2023.

\bibitem{gong2019efficient}
Linyuan Gong, Di~He, Zhuohan Li, Tao Qin, Liwei Wang, and Tieyan Liu.
\newblock Efficient training of bert by progressively stacking.
\newblock In {\em ICML}, pages 2337--2346. PMLR, 2019.

\bibitem{yang2023global}
Huanrui Yang, Hongxu Yin, Maying Shen, Pavlo Molchanov, Hai Li, and Jan Kautz.
\newblock Global vision transformer pruning with hessian-aware saliency.
\newblock In {\em CVPR}, pages 18547--18557, 2023.

\bibitem{yu2022width}
Fang Yu, Kun Huang, Meng Wang, Yuan Cheng, Wei Chu, and Li~Cui.
\newblock Width \& depth pruning for vision transformers.
\newblock In {\em AAAI}, volume~36, pages 3143--3151, 2022.

\bibitem{lagunas2021block}
Fran{\c{c}}ois Lagunas, Ella Charlaix, Victor Sanh, and Alexander~M Rush.
\newblock Block pruning for faster transformers.
\newblock In {\em EMNLP}, pages 10619--10629, 2021.

\bibitem{xia2022structured}
Mengzhou Xia, Zexuan Zhong, and Danqi Chen.
\newblock Structured pruning learns compact and accurate models.
\newblock In {\em ACL}, pages 1513--1528, 2022.

\bibitem{rao2021dynamicvit}
Yongming Rao, Wenliang Zhao, Benlin Liu, Jiwen Lu, Jie Zhou, and Cho-Jui Hsieh.
\newblock Dynamicvit: Efficient vision transformers with dynamic token sparsification.
\newblock {\em NeurIPS}, 34:13937--13949, 2021.

\bibitem{meng2022adavit}
Lingchen Meng, Hengduo Li, Bor-Chun Chen, Shiyi Lan, Zuxuan Wu, Yu-Gang Jiang, and Ser-Nam Lim.
\newblock Adavit: Adaptive vision transformers for efficient image recognition.
\newblock In {\em CVPR}, pages 12309--12318, 2022.

\bibitem{fayyaz2022adaptive}
Mohsen Fayyaz, Soroush~Abbasi Koohpayegani, Farnoush~Rezaei Jafari, Sunando Sengupta, Hamid Reza~Vaezi Joze, Eric Sommerlade, Hamed Pirsiavash, and J{\"u}rgen Gall.
\newblock Adaptive token sampling for efficient vision transformers.
\newblock In {\em ECCV}, pages 396--414. Springer, 2022.

\bibitem{kong2022spvit}
Zhenglun Kong, Peiyan Dong, Xiaolong Ma, Xin Meng, Wei Niu, Mengshu Sun, Xuan Shen, Geng Yuan, Bin Ren, Hao Tang, et~al.
\newblock Spvit: Enabling faster vision transformers via latency-aware soft token pruning.
\newblock In {\em ECCV}, pages 620--640. Springer, 2022.

\bibitem{yin2022vit}
Hongxu Yin, Arash Vahdat, Jose~M Alvarez, Arun Mallya, Jan Kautz, and Pavlo Molchanov.
\newblock A-vit: Adaptive tokens for efficient vision transformer.
\newblock In {\em CVPR}, pages 10809--10818, 2022.

\bibitem{xu2023q}
Sheng Xu, Yanjing Li, Mingbao Lin, Peng Gao, Guodong Guo, Jinhu L{\"u}, and Baochang Zhang.
\newblock Q-detr: An efficient low-bit quantized detection transformer.
\newblock In {\em CVPR}, pages 3842--3851, 2023.

\bibitem{li2022q}
Yanjing Li, Sheng Xu, Baochang Zhang, Xianbin Cao, Peng Gao, and Guodong Guo.
\newblock Q-vit: Accurate and fully quantized low-bit vision transformer.
\newblock {\em NeurIPS}, 35:34451--34463, 2022.

\bibitem{he2023bivit}
Yefei He, Zhenyu Lou, Luoming Zhang, Jing Liu, Weijia Wu, Hong Zhou, and Bohan Zhuang.
\newblock Bivit: Extremely compressed binary vision transformers.
\newblock In {\em ICCV}, pages 5651--5663, 2023.

\bibitem{le2023binaryvit}
Phuoc-Hoan~Charles Le and Xinlin Li.
\newblock Binaryvit: Pushing binary vision transformers towards convolutional models.
\newblock In {\em CVPR}, pages 4664--4673, 2023.

\bibitem{chen2022bert2bert}
Cheng Chen, Yichun Yin, Lifeng Shang, Xin Jiang, Yujia Qin, Fengyu Wang, Zhi Wang, Xiao Chen, Zhiyuan Liu, and Qun Liu.
\newblock bert2bert: Towards reusable pretrained language models.
\newblock In {\em ACL}, pages 2134--2148, 2022.

\bibitem{yuan2020growing}
Xin Yuan, Pedro Savarese, and Michael Maire.
\newblock Growing efficient deep networks by structured continuous sparsification.
\newblock In {\em ICLR}, 2021.

\bibitem{wen2020autogrow}
Wei Wen, Feng Yan, Yiran Chen, and Hai Li.
\newblock Autogrow: Automatic layer growing in deep convolutional networks.
\newblock In {\em KDD}, pages 833--841, 2020.

\bibitem{wang2023efficienttrain}
Yulin Wang, Yang Yue, Rui Lu, Tianjiao Liu, Zhao Zhong, Shiji Song, and Gao Huang.
\newblock Efficienttrain: Exploring generalized curriculum learning for training visual backbones.
\newblock In {\em ICCV}, pages 5852--5864, 2023.

\bibitem{li2022automated}
Changlin Li, Bohan Zhuang, Guangrun Wang, Xiaodan Liang, Xiaojun Chang, and Yi~Yang.
\newblock Automated progressive learning for efficient training of vision transformers.
\newblock In {\em CVPR}, pages 12486--12496, 2022.

\bibitem{pan2022budgeted}
Xuran Pan, Xuan Jin, Yuan He, Shiji Song, Gao Huang, et~al.
\newblock Budgeted training for vision transformer.
\newblock In {\em ICLR}, 2022.

\bibitem{lee2022deduplicating}
Katherine Lee, Daphne Ippolito, Andrew Nystrom, Chiyuan Zhang, Douglas Eck, Chris Callison-Burch, and Nicholas Carlini.
\newblock Deduplicating training data makes language models better.
\newblock In {\em ACL}, pages 8424--8445, 2022.

\bibitem{tan2021efficientnetv2}
Mingxing Tan and Quoc Le.
\newblock Efficientnetv2: Smaller models and faster training.
\newblock In {\em ICLR}, pages 10096--10106. PMLR, 2021.

\bibitem{mcdanel2022accelerating}
Bradley McDanel and Chi~Phuong Huynh.
\newblock Accelerating vision transformer training via a patch sampling schedule.
\newblock {\em arXiv preprint arXiv:2208.09520}, 2022.

\bibitem{shen2023efficient}
Li~Shen, Yan Sun, Zhiyuan Yu, Liang Ding, Xinmei Tian, and Dacheng Tao.
\newblock On efficient training of large-scale deep learning models: A literature review.
\newblock {\em arXiv preprint arXiv:2304.03589}, 2023.

\bibitem{yang2023efficient}
Yuedong Yang, Guihong Li, and Radu Marculescu.
\newblock Efficient on-device training via gradient filtering.
\newblock In {\em CVPR}, pages 3811--3820, 2023.

\bibitem{ye2020accelerating}
Xucheng Ye, Pengcheng Dai, Junyu Luo, Xin Guo, Yingjie Qi, Jianlei Yang, and Yiran Chen.
\newblock Accelerating cnn training by pruning activation gradients.
\newblock In {\em ECCV}, pages 322--338. Springer, 2020.

\bibitem{fu2020fractrain}
Yonggan Fu, Haoran You, Yang Zhao, Yue Wang, Chaojian Li, Kailash Gopalakrishnan, Zhangyang Wang, and Yingyan Lin.
\newblock Fractrain: Fractionally squeezing bit savings both temporally and spatially for efficient dnn training.
\newblock {\em NeurIPS}, 33:12127--12139, 2020.

\bibitem{wang2019e2}
Yue Wang, Ziyu Jiang, Xiaohan Chen, Pengfei Xu, Yang Zhao, Yingyan Lin, and Zhangyang Wang.
\newblock E2-train: Training state-of-the-art cnns with over 80\% energy savings.
\newblock {\em NeurIPS}, 32, 2019.

\bibitem{li2019budgeted}
Mengtian Li, Ersin Yumer, and Deva Ramanan.
\newblock Budgeted training: Rethinking deep neural network training under resource constraints.
\newblock In {\em ICLR}, 2019.

\bibitem{zhang2019autoassist}
Jiong Zhang, Hsiang-Fu Yu, and Inderjit~S Dhillon.
\newblock Autoassist: A framework to accelerate training of deep neural networks.
\newblock {\em NeurIPS}, 32, 2019.

\bibitem{epoch2021backwardforwardFLOPratio}
Marius Hobbhahn and Jaime Sevilla.
\newblock What’s the backward-forward flop ratio for neural networks?
\newblock \url{https://epochai.org/blog/backward-forward-FLOP-ratio}, 2021.
\newblock Accessed: 2023-9-28.

\bibitem{van2008visualizing}
Laurens van~der Maaten and Geoffrey Hinton.
\newblock Visualizing data using t-sne.
\newblock {\em JMLR}, 9:2579--2605, 2008.

\bibitem{deng2009imagenet}
Jia Deng, Wei Dong, Richard Socher, Li-Jia Li, Kai Li, and Li~Fei-Fei.
\newblock Imagenet: A large-scale hierarchical image database.
\newblock In {\em CVPR}, pages 248--255. Ieee, 2009.

\bibitem{krizhevsky2009learning}
Alex Krizhevsky et~al.
\newblock Learning multiple layers of features from tiny images.
\newblock 2009.

\bibitem{paszke2019pytorch}
Adam Paszke, Sam Gross, Francisco Massa, Adam Lerer, James Bradbury, Gregory Chanan, Trevor Killeen, Zeming Lin, Natalia Gimelshein, Luca Antiga, et~al.
\newblock Pytorch: An imperative style, high-performance deep learning library.
\newblock {\em NeurIPS}, 32, 2019.

\bibitem{raghu2021vision}
Maithra Raghu, Thomas Unterthiner, Simon Kornblith, Chiyuan Zhang, and Alexey Dosovitskiy.
\newblock Do vision transformers see like convolutional neural networks?
\newblock {\em NeurIPS}, 34:12116--12128, 2021.

\bibitem{fang2021you}
Yuxin Fang, Bencheng Liao, Xinggang Wang, Jiemin Fang, Jiyang Qi, Rui Wu, Jianwei Niu, and Wenyu Liu.
\newblock You only look at one sequence: Rethinking transformer in vision through object detection.
\newblock {\em NeurIPS}, 34:26183--26197, 2021.

\bibitem{lin2014microsoft}
Tsung-Yi Lin, Michael Maire, Serge Belongie, James Hays, Pietro Perona, Deva Ramanan, Piotr Doll{\'a}r, and C~Lawrence Zitnick.
\newblock Microsoft coco: Common objects in context.
\newblock In {\em ECCV}, pages 740--755. Springer, 2014.

\end{thebibliography}
}

\ifarxiv \clearpage \appendix \section{Implementation Details}

\subsection{Details of Applying \OurMethod to DeiT and LV-ViT}

\begin{figure}[htbp]
    \centering
    \includegraphics[scale = 0.3]{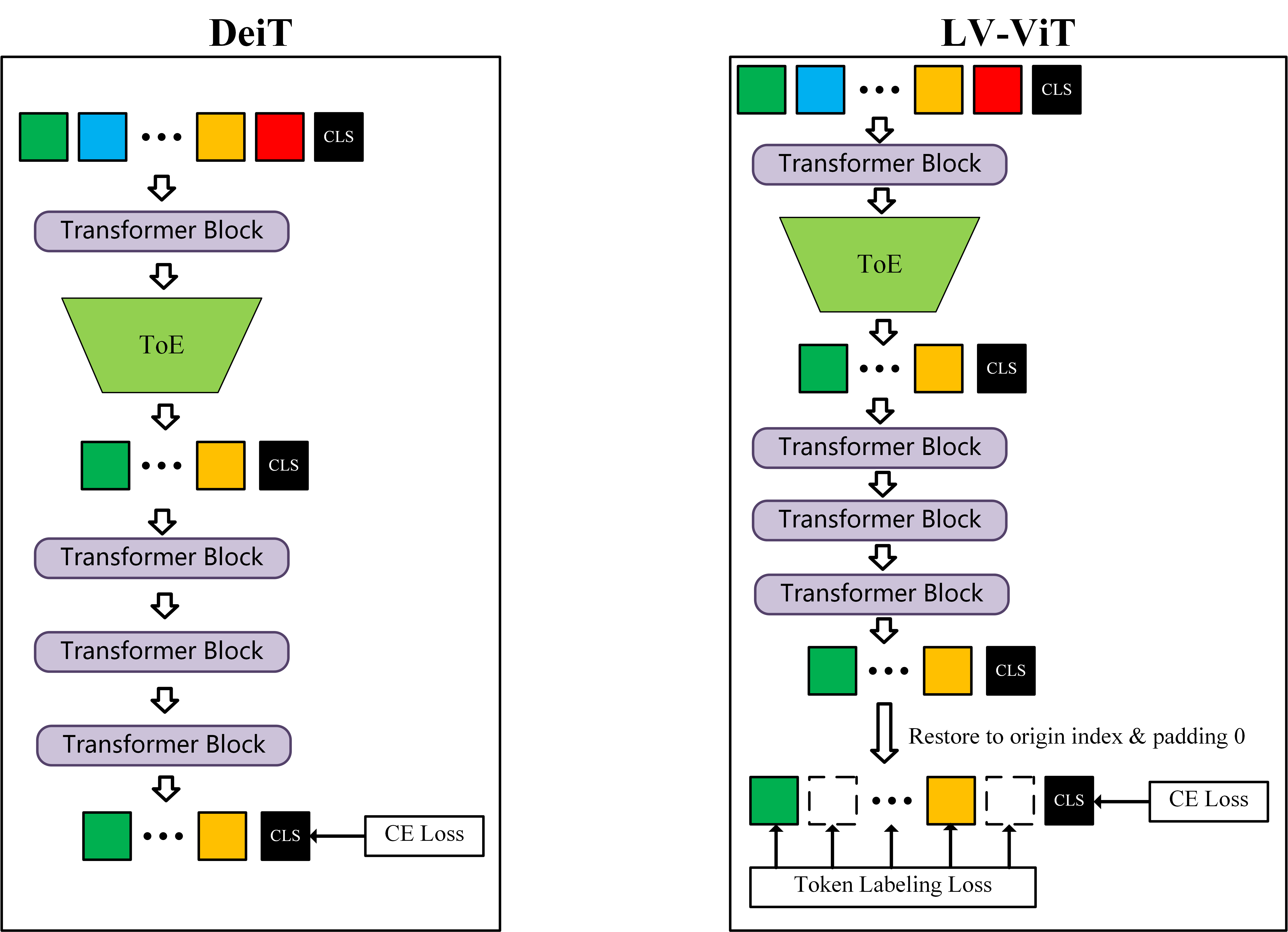}
    \caption{Details of applying \OurMethod to DeiT and LV-ViT during training. Dotted cubes denote the tokens are all-zero vectors.}
    \label{sup_fig:sup_deit_and_lvvit_detail}
\end{figure}

For DeiT~\cite{touvron2021training} and LV-ViT~\cite{jiang2021all}, we apply \OurMethod to the output tokens of the first block.
For the training of DeiT, we simply reduce the tokens by \OurMethod.
But for LV-ViT requiring the token index, we employ \textit{zero-padding} on the reduced output tokens of last Transformer block and restore the tokens to their original index.
The details are presents in Fig.~\ref{sup_fig:sup_deit_and_lvvit_detail}. We also use the same way (by adding ToE at the first block output tokens) to combine our ToE with EfficientTrain to achieve the better performance, which is summarized in Tab. 4 of the main paper.

\subsection{Details of Breaking the Restriction of Hyper-parameter Consistency}

Firstly, for the training of DeiT, we follow the hyper-parameters of original paper~\cite{touvron2021training}. We set the batch size to be $1,024$, learning rate to be $1e-3$ using a cosine scheduler with warmup, and the decay to the minimal learning rate of $1e-5$. 
We employ the AdamW optimizer, whose weight decay is set to $5e-2$.

In Tab. 2 of the main paper, we relax the restriction of hyper-parameter consistency to achieve better results. We will decribe the following training details for the \OurMethod$_{0.4}^{\mathrm{Hyper}}$ and \OurMethod$_{0.5}^{\mathrm{Hyper}}$.
In fact, we only change the minimal learning rate and use the more elaborate training schedule.
Specifically, we set minimal learning rate to $2e-4$ and change default training schedule of \OurMethod from [0$\rightarrow$100, 101$\rightarrow$200, 201$\rightarrow$300] for three stages with default average splitting epochs to [0$\rightarrow$130, 131$\rightarrow$260, 261$\rightarrow$300]. %

\subsection{Training Details of Fine-tuning}

Following the fine-tuning process in~\cite{touvron2021training}, we fine-tune DeiT for $1,000$ epochs with an initial learning rate of $3e-5$, and the batch size of $768$ per GPU for four GPUs on CIFAR-10/100~\cite{krizhevsky2009learning}~\footnote{https://github.com/facebookresearch/dino/issues/144}~\footnote{https://github.com/facebookresearch/deit/issues/45}.
The input image size of $32\times32$ are resized to $224\times224$.
Other hyper-parameters and strategies are the same as the pre-training process on ImageNet-1K~\cite{deng2009imagenet}.

\subsection{Details of Training time}

\begin{table}[t]
    \caption{Each epoch training time in the different training stages.}
    \centering
    \resizebox{0.85\linewidth}{!}{
        \begin{tabular}{c|c|c|c}
            \toprule
            \multirow{3}{*}{Model}              & \multicolumn{3}{c}{Training time}                                      \\
                                                & \multicolumn{3}{c}{(GPU hours per training epoch)}                     \\
            \cmidrule{2-4}
                                                & Stage-1                                            & Stage-2 & Stage-3 \\
            \midrule
            DeiT-tiny + \OurMethod$_{r_1=0.5}$  & 395s                                               & 542s    & 655s    \\
            DeiT-small + \OurMethod$_{r_1=0.5}$ & 948s                                               & 1,244s   & 1,488s   \\
            DeiT-base + \OurMethod$_{r_1=0.5}$  & 2,028s                                              & 2,784s   & 3,512s   \\
            DeiT-base + \OurMethod$_{r_1=0.4}$  & 1,852s                                              & 2,740s   & 3,512s   \\
            LV-ViT-T + \OurMethod$_{r_1=0.4}$   & 1,180s                                              & 1,356s   & 1,566s   \\
            LV-ViT-S + \OurMethod$_{r_1=0.4}$   & 1,828s                                              & 2,356s   & 2,848s   \\
            LV-ViT-M + \OurMethod$_{r_1=0.4}$   & 2,596s                                                   &  3,512s       &  4,424s       \\
            \bottomrule
        \end{tabular}

    }
    \label{sup_tab:sup_training_time}
\end{table}

\begin{figure*}[htbp]
    \centering
    \includegraphics[scale = 0.42]{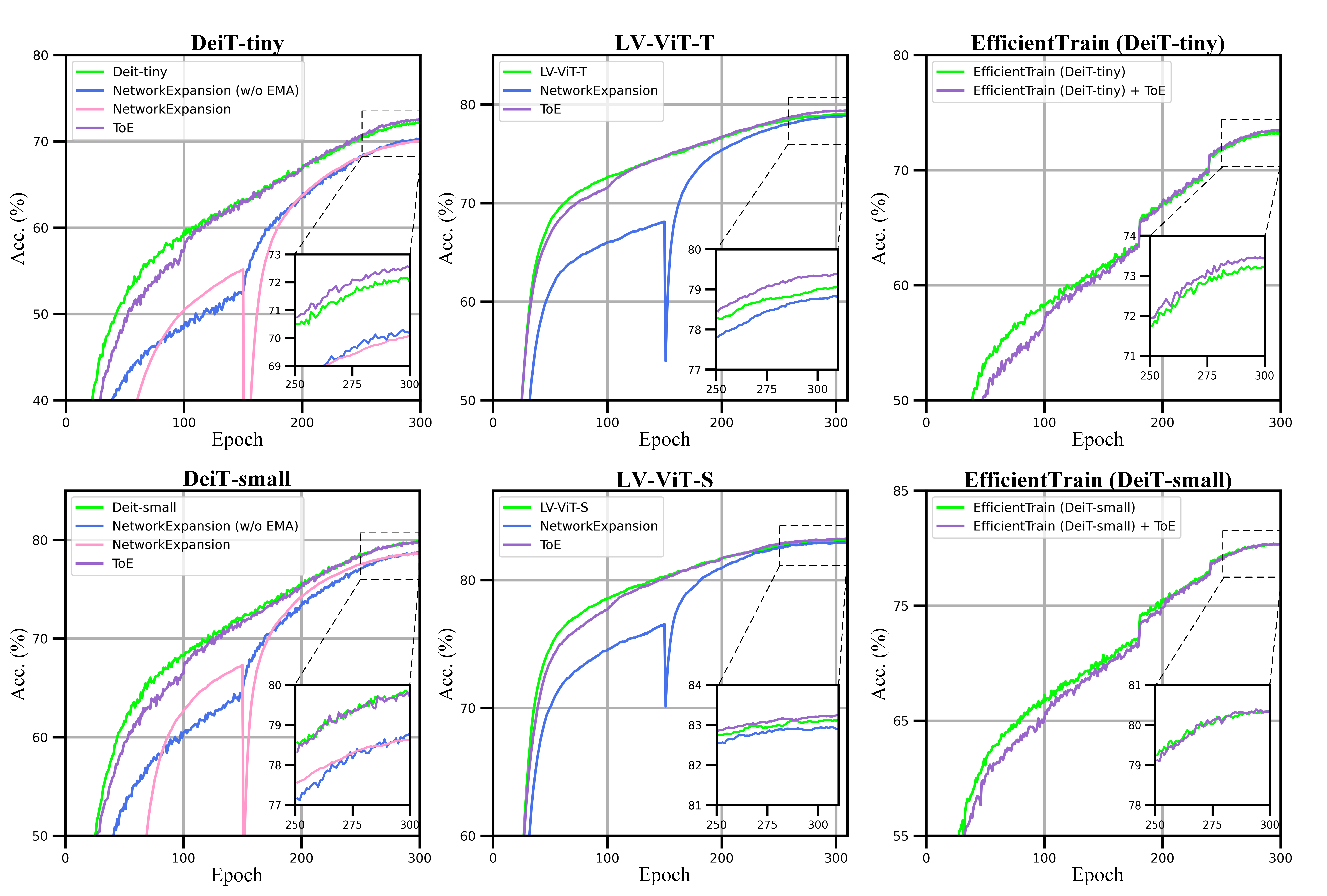}
    \caption{Validation Top-1 accuracy of DeiT-tiny\&small and LV-ViT-T\&S on ImageNet-1k during training with different methods. DeiT does \textit{not} use the EMA strategy by default, while LV-ViT uses the EMA strategy by default.}
    \label{sup_fig:sup_deit_and_lvvit_acc}
\end{figure*}

\begin{figure*}[htbp]
    \centering
    \includegraphics[scale = 0.3]{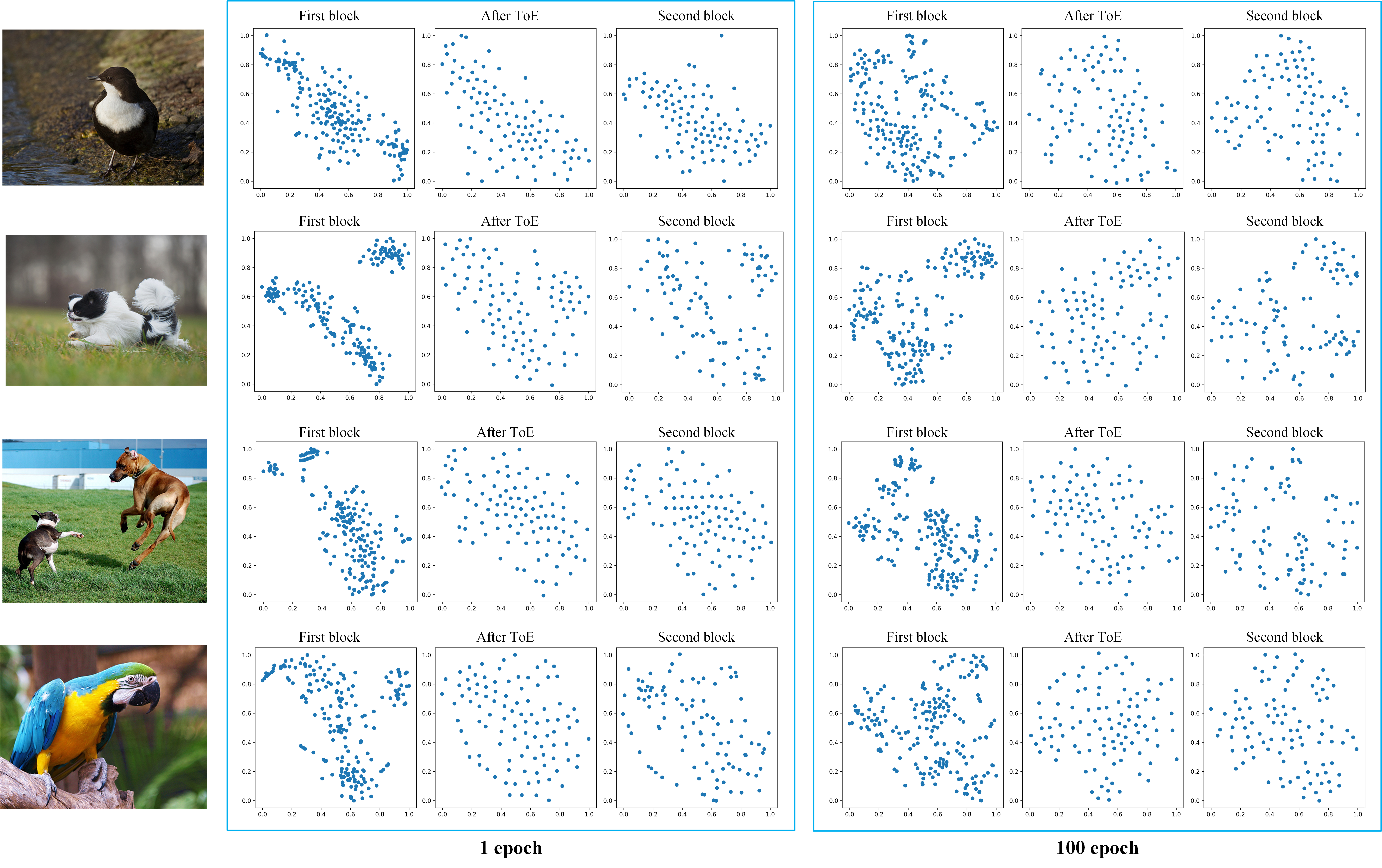}
    \caption{More visualization for the feature distribution of token set. Continuation of Fig. 2 of the main paper.}
    \label{sup_fig:sup_t_sne}
\end{figure*}

The detailed training time per training epoch for applying \OurMethod to the models in different training stages are presented in Tab.~\ref{sup_tab:sup_training_time}.
The training time is averagely measured by 3 times running.

\section{Additional Results}

\begin{table}[t]
\footnotesize
\centering
\captionof{table}{Results of ToE on YOLOS for COCO object detection. We use eight NVIDIA RTX A6000 GPUs with 150 epochs for YOLOS-S.}
    \resizebox{0.8\linewidth}{!}{
        \begin{tabular}[t]{c|c|cc}
            \toprule
           Model  &   Method     & AP  & Total GPU hours  \\
           \midrule
       \multirow{2}{*}{YOLOS-S}     &   Baseline~\cite{fang2021you}                  & 36.1     &  1,193h    \\
         &   \cellcolor[gray]{0.8}ToE$_{r_1=0.5}$ (Ours)      & \cellcolor[gray]{0.8}36.0 (-0.1)   & \cellcolor[gray]{0.8}964h (1.24$\times$) \\
            \bottomrule
        \end{tabular}
    }
    \label{sup_tab:sup_yolos_coco}
\end{table}

\subsection{Additional Results for Object Detection}

In Tab.~\ref{sup_tab:sup_yolos_coco}, \OurMethod applied into YOLOS~\cite{fang2021you}.
\OurMethod reduces $229$ hours with 1.24$\times$ training speedup for training YOLOS-S on COCO~\cite{lin2014microsoft} with the only $0.1$ AP drop.

\subsection{More Validation Curves of Training Process}

We present the validation curves of training process for integrating into \OurMethod to DeiT, LV-ViT and EfficientTrain framework~\cite{wang2023efficienttrain} in Fig.~\ref{sup_fig:sup_deit_and_lvvit_acc}.
For the different ViTs and efficient training frameworks, \OurMethod can general accelerate the training process in a lossless manner.

\subsection{Visualization of \OurMethod}

More visualizations of \OurMethod as a continuation of Fig. 2 of the main paper are presented in Fig.~\ref{sup_fig:sup_t_sne}.
\OurMethod preserves the distribution integrity of intermediate features in the original token set.

 \fi

\end{document}